\documentclass[acmsmall]{acmart}

\usepackage{booktabs} 
\usepackage{hyperref}   
\usepackage{url}  

\usepackage{multirow}  
\usepackage{color}  
\newtheorem{property}{Property}    

\usepackage{diagbox}
\usepackage{amsmath}
\usepackage{arydshln} 
\usepackage{color} 

\usepackage[linesnumbered,ruled]{algorithm2e} 

\SetAlFnt{\small}
\SetAlCapFnt{\small}
\SetAlCapNameFnt{\small}
\SetAlCapHSkip{0pt}
\IncMargin{-\parindent}

\acmJournal{TMIS}

\setcopyright{rightsretained}

\begin{document}
\title{Smart System: Joint Utility and Frequency for Pattern Classification}

\author{Qi Lin*}
\affiliation{
	\institution{Jinan University of Birmingham Joint Institute}
	\city{Guangzhou}
	\country{China}
}
\email{qlin0910@gmail.com}

\author{Wensheng Gan}
\authornote{Both authors contribute equally to the paper.}
\affiliation{
	\institution{Jinan University}
	\city{Guangzhou}
	\country{China}
}
\email{wsgan001@gmail.com}

\author{Yongdong Wu}
\affiliation{
	\institution{Jinan University}
	\city{Guangzhou}
	\country{China}
}
\email{wuyd007@qq.com}

\author{Jiahui Chen}
\affiliation{%
	\institution{Guangdong University of Technology}
	\city{Guangzhou}
\country{China}
}
\email{csjhchen@gmail.com}

\author{Chien-Ming Chen}
\authornote{This is the corresponding author}
\affiliation{%
	\institution{Shandong University of Science and Technology}
	\city{Qingdao}
	\country{China}
}
\email{chienmingchen@ieee.org}

\begin{abstract}

Nowadays, the environments of smart systems for Industry 4.0 and Internet of Things (IoT) are experiencing fast industrial upgrading. Big data technologies such as design making, event detection, and classification are developed to help manufacturing organizations to achieve smart systems. By applying data analysis, the potential values of rich data can be maximized and thus help manufacturing organizations to finish another round of upgrading. In this paper, we propose two new algorithms with respect to big data analysis, namely UFC$_{gen}$ and UFC$_{fast}$. Both algorithms are designed to collect three types of patterns to help people determine the market positions for different product combinations. We compare these algorithms on various types of datasets, both real and synthetic. The experimental results show that both algorithms can successfully achieve pattern classification by utilizing three different types of interesting patterns from all candidate patterns based on user-specified thresholds of utility and frequency. Furthermore, the list-based UFC$_{fast}$ algorithm outperforms the level-wise-based UFC$_{gen}$ algorithm in terms of both execution time and memory consumption. 

\end{abstract}


%
%



\maketitle

\renewcommand{\shortauthors}{Qi Lin \textit{et al.}}

\section{Introduction}

In today's manufacturing environment, massive raw data are collected from shops, online e-commerce platforms or through sensors and electronic equipment. These data can be related to user behaviors, transaction records, the characteristics of products, production lines and so on. The size of data is increasingly grown up in the manufacturing industries \cite{choudhary2009data,dogan2020machine} and this has led to a very important issue, that is how to discover the potential knowledge from the high-volume databases. During the past few years, big data analysis has been attached great importance to different fields, especially for smart manufacturing. In big data analysis, knowledge discovery in databases (KDD) \cite{Chen1996Data} aims to extract useful patterns from data, and data mining (DM) \cite{Chen1996Data,gan2017data} is the key step in KDD since it provides the algorithm or model to efficiently discover the useful patterns and knowledge. Association rule mining (ARM) \cite{agrawal1994fast, hong1999mining} and pattern classification are two of the most attractive and studied fields in data mining. ARM has been very popular in the past two decades. Scientists have intended to determine the correlation and association for various items, thereby predicting future trends or developing feasible strategies. To describe the frequency of a pattern and the reliability of a rule, the concepts of support and confidence are adopted to record such information in a database. For ARM, the first step is to filter data, a common and traditional method is to formulate rules by utilizing algorithms in high-frequency itemset mining (HFIM)  \cite{luna2019frequent,hong1999mining,han2004mining}. During this step, high-frequency itemsets (HFIs) are discovered, and the other itemsets that do not meet the minimum frequency threshold are eliminated. The second step is to determine the hidden rules or patterns in HFIM and choose the rules with higher confidence. Finally, these interesting rules are applied in the real world to help design appropriate strategies for different tasks. In the past, a number of ARM approaches have been proposed, such as Apriori \cite{han2011data}, FP-growth \cite{han2004mining}, H-Mine \cite{pei2001h}, and Eclat \cite{zaki2000scalable}.

Based on ARM, pattern classification (PC) and classification based on associations (CBA) \cite{thabtah2007review, abdelhamid2014associative,nguyen2012classification,thabtah2004mmac} have been studied to label items or build classifiers to predict future trends. It makes use of the rule discovery process in ARM by extracting efficient rules which can precisely generalize the training databases. Nowadays, pattern classification has been applied in various world application such as commercial prediction, finance analysis, phishing website detection \cite{abdelhamid2014phishing}. For example, algorithms of pattern classification extract classifiers containing association rules of high confidence to put different labels on websites, thereby identifying which website can be viewed as a phishing website. As ARM normally adopts frequency as the sole standard to extract valuable knowledge, every item is assigned to the same utility (value), and therefore frequency becomes the only precondition to obtain the association rules. It is noticeable that such a strategy ignores the variety of importance in items, and strategies developed to address this may not improve the sales of some items. To overcome this drawback, utility-driven pattern mining \cite{gan2021survey} has been proposed. In smart manufacturing, the high-utility patterns are always welcomed by merchants, because they simply create higher profits. High-utility pattern mining (HUPM) \cite{chan2003mining,liu2005two,lin2017fdhup} assigns different utilities to items and considers both occurred quantity (internal utility) and utility (external utility). In recent years, a series of alternative approaches have been proposed, including IHUP \cite{ahmed2009efficient}, HUI-Miner \cite{liu2012mining}, UMEpi \cite{gan2019utility}, UP-growth \cite{tseng2010up}, TKU \cite{tseng2015efficient}, and so on.

In the actual process of manufacturing, both frequency and utility are the essential indicators for determining the position of a product combination. On the one hand, in HFIM, high frequency itemsets are discovered, and they are often related to the ARM. A number of association rules are conducted from these frequent itemsets. However, if a majority of these itemsets have low utility, then it might be a waste of resource  unless these itemsets have a high frequency so that new strategies can be applied. On the other hand, we do not expect similar situation in HUPM. In manufacturing, if many itemsets with high utility being discovered are quite infrequent, then the discovered rules may be less valuable. It is clear that one of the major challenges in pattern mining for manufacturing is how to better locate each pattern's position and design more specific and valuable strategies for bundling sale. Therefore, if frequency and utility can be taken into account in big data analysis, then we can better evaluate each combination of different products, and then obtain a more accurate position of each combination, thus develop more effective and specific bundling sale strategies. In this study, we introduce a new model for pattern classification. We formulate this model as joint utility and frequency for pattern classification, and also develop two effective algorithms.

Based on the properties of frequency and utility, there are four types of patterns in pattern classification, including (i) High Frequency and High Utility Itemset (HFHUI); (ii) High Frequency and Low Utility Itemset (HFLUI); (iii) Low Frequency and High Utility Itemset (LFHUI); and (iv) Low Frequency and Low Utility Itemset (LFLUI). Since LFLUI is meaningless and numerous, our proposed model will focus on the interesting three types, i.e., HFHUI, HFLUI, and LFHUI. The key contributions of this paper are summarized as follows:

\begin{itemize}	
	
	\item In data analysis, only a few studies consider both frequency and utility as two common standards to discover interesting patterns. In this study, two algorithms namely UFC$_{gen}$ and UFC$_{fast}$ are developed to help better locate the market locations for different product combinations in real-world applications, such as smart manufacturing. Both algorithms can efficiently collect three different types of patterns.

	\item Based on utility and frequency, several powerful strategies are developed to prune the search space. Different strategies are applied in UFC$_{gen}$ and UFC$_{fast}$ to ensure the reliability of pattern classification.
	
	\item A special novel structure named frequency-utility-list (FU-list for short) is developed in the efficient UFC$_{fast}$ algorithm. Such a structure stores information about transaction identifier, frequency and remaining utility of patterns.
	
	\item Extensive experiments on real and synthetic datasets show that the UFC$_{gen}$ and UFC$_{fast}$ algorithms can be applied to various sizes of datasets, and UFC$_{fast}$ has a remarkable performance than UFC$_{gen}$.
\end{itemize}

Note that some key concepts and an initial algorithm were presented in a preliminary version \cite{lin2021joint} of this article. The remainder of this extended version is organized as follows. The relevant literature and necessary background of association rule mining, associative classification, and utility-driven pattern mining are respectively reviewed in Section \ref{relatedwork}. Some descriptions of associative classification and the statement of the studied problem are provided and discussed in Section \ref{background}. The detailed process of level-based algorithm are presented in Section \ref{algorithm1}. Then, in Section \ref{algorithm2} we provide a supplementary algorithm, which has a better performance on solving the pattern classification problem. Extensive experimental results are presented in Section \ref{experiments}. Finally, Section \ref{conclusion} concludes the paper and highlights several future studies.

\section{Related Work}
\label{relatedwork}

In this section, we review the related works which have been studied in big data analysis, including data mining in manufacturing, utility-driven pattern mining, and pattern classification.

\subsection{Data mining in manufacturing}

The general goal of data mining is to find the potential relationship between items, hidden pattern and predict the future trend. In smart manufacturing, the algorithm for data mining can help diagnose fault information, discover uncovering relationship among products, and design complex strategies. With the extensive use of machines and a large production of items, data mining in manufacturing usually plays an increasingly important role. In the past two decades, a number of approaches have been developed and applied in the real process of manufacturing \cite{choudhary2009data,dogan2020machine}. For instance, Fan \textit{et al.} \cite{fan2001data} proposed an algorithm that can automatically discover knowledge from production databases for fault diagnosis, and finally it can optimize device performance with a given target.  In 2017, Nakata \textit{et al.} \cite{nakata2017comprehensive} proposed a system that can identity the cause of failure from wafer failure map patterns and manufacturing histories. Then, by applying big data analysis into different stages, the system can help engineers to support their work.

ARM \cite{agrawal1993mining,brin1997beyond} was first proposed to analyze basket data and is nowadays widely applied in different aspects. In smart manufacturing industries, ARM can be used to find the association from products and induce a number of rules that can help predict user behavior. In general, there are two main steps in association rule mining. The first step is to discover frequent itemsets by calculating the occurrence of each possible combination in the datasets. And in the next step, it simply generates and selects rules from the itemset by acknowledging the confidence of each rule. In the past, approaches such as Apriori \cite{agrawal1994fast} and FP-growth \cite{han2004mining} were applied to obtain the HFIs. In these approaches, Apriori proposed by Agrawal and Srikant \cite{agrawal1994fast} in 1994 is the most classic algorithm. According to the frequency downward closure property, if a 1-itemset's frequency cannot meet the minimum frequency (\textit{min\_fre}) threshold, then all of its supersets cannot be a high-frequency itemset. Apriori adopts this property to discover the HFIs using a generate-and-test search. Although it can avoid a brute-force manner, it usually needs to scan the database multiple times. Therefore, it usually requires high consumption of execution time and a large search space. To overcome these shortages, the FP-growth algorithm  \cite{han2004mining} only requires twice scans of the database by utilizing a tree structure, called FP-tree, that extends its branches by using the prefix. Compared with Apriori, the FP-tree algorithm has less time and space complexity. To date, a large number of HFIM and ARM algorithms have been proposed in different domains \cite{xie2010max,gan2017data,zaki2000scalable,luna2019frequent}.

\subsection{Utility-driven pattern mining}

In the field of ARM, a number of rules with high support and confidence can be discovered by applying different existing algorithms, but the items included in the rules may not be the best target for the company. Generally, the profit of products is emphasized in smart manufacturing, people are eager to realize which product combination can bring more profits. High-utility pattern mining (HUPM) \cite{gan2021survey,2gan2018survey,gan2018privacy} can well solve the problem, since for HUPM in manufacturing organizations, utility can often be replaced by profit. As the frequency downward closure property was first developed in HFIM, to maintain the similar property in HUPM, the transaction-weighted utility downward closure property was first applied in a two-phase model that proposed by Liu \textit{et al.} \cite{liu2005two}. In the first phase, it scans the database level-by-level to obtain a set of candidate patterns whose TWU satisfies the minimum utility threshold. In the second phase, the database is scanned again to compute the specific utility for each candidate. Therefore, all the actual HUIs are found in the second phase. To date, a series of utility-driven mining approaches have been proposed, as reviewed in \cite{gan2021survey}, including IHUP \cite{ahmed2009efficient} for processing dynamic data, UP-growth \cite{tseng2010up} for processing static data, utility mining in uncertain data \cite{lin2016efficient,gan2020utility}, FHN \cite{lin2016fhn} for processing transaction data containing both positive and negative utility values, UMEpi \cite{gan2019utility} for discovering high-utility episodes, up-to-date high-utility pattern mining \cite{lin2015efficient,gan2020utility2}, TKU \cite{tseng2015efficient} and TKUS \cite{zhang2021tkus} for top-$k$ mining, HUPM in dynamic profit databases \cite{nguyen2019mining} or noisy databases \cite{baek2021approximate}, and utility mining on sequence data \cite{gan2020proum,gan2021fast,zhang2021shelf}. Besides, based on the new utility-occupancy concept, the HUOPM \cite{gan2020huopm} and UHUOPM \cite{chen2021discovering} algorithms are proposed. Some researches tried to conduct a linear combination of frequency and utility as a weighted model, and thus more information about the patterns can be included \cite{wang2007pushing,gan2018extracting}. Similarly, Shao \textit{et al.} \cite{shao2015mining} proposed an algorithm to mine a combined pattern with high utility and frequency. Such a pattern considers the utility in generated associative rules, which aims to discover rules containing HUIs. In the past, a fast utility frequent mining algorithm (FUFM) \cite{shankar2009fast} was introduced.

\subsection{Pattern classification}

In smart manufacturing, pattern classification aims to automatically classify each pattern (a single product or a product combination) as to help decision makers to locate market position of each pattern and design more suitable strategies. To be different from general pattern mining applying single goal, pattern classification often consider multiple goals. For example, some researches tried to classify patterns according to frequency and utility together. In the past, some researchers tried conducting a linear combination of frequency and utility as a weighted model, and thus more information about the patterns can be included. Lin \textit{et al.} \cite{lin2017fdhup} introduced the concept of discriminative high-utility pattern with strong frequency affinity. Similarly, Shao \textit{et al.} \cite{shao2015mining} proposed an algorithm to mine a combined pattern with high utility and frequency. Such a pattern considers the utility in generated associative rules, aiming to discover rules containing high-utility patterns. Meanwhile, in the past, a fast utility frequent mining algorithm (FUFM) \cite{shankar2009fast} was also introduced, which conducted two approaches about FRM and HUPM in two different phases in order to categorize the pattern. The jointly utility-frequency approach discussed in the following sections utilizes both pattern classification and utility-driven pattern mining. Besides, the two designed algorithms are quite different. The first one adopts a two-phase manner, and the second one utilizes a vertical data structure containing transaction number, frequency, and remaining utility. Both algorithms aim to collect three different kinds of patterns based on frequency and utility.

\section{Preliminaries} \label{background}

\subsection{Definitions}

In this subsection, the preliminaries and definitions of key terms related to pattern classification and utility-driven mining are presented. A transaction database with the utility-table as a running example is given in Tables \ref{table:db} and \ref{table:utable}, respectively.

\begin{table}[h]
	\centering
	\caption{Example database}
	\begin{tabular}{ccc}
		\hline
		\textbf{\textit{tid}}  & \textbf{Transaction} & \textbf{Utility}  \\
		\hline
		$T_1$ & ($A$, 1), ($B$, 2), ($C$, 1)&   \$13     \\
		$T_2$ & ($A$, 2), ($B$, 3), ($F$, 2)&   \$23        \\
		$T_3$ & ($B$, 2), ($D$, 2), ($E$, 2)&   \$16      \\
		$T_4$ & ($C$, 2), ($D$, 1), ($F$, 1), ($G$, 3) &  \$10   \\
		$T_5$ & ($B$, 1), ($C$, 2), ($F$, 2), ($G$, 1)&  \$12 \\
		\hline
	\end{tabular}
	\label{table:db}
\end{table}

\begin{table}[h]
	\centering
	\caption{Utility table}
	\begin{tabular}{lccccccc}
		\hline
		\textbf{Item} &\textbf{\emph{A}} & \textbf{\emph{B}} & \textbf{\emph{C}}& \textbf{\emph{D}}& \textbf{\emph{E}}& \textbf{\emph{F}}&\textbf{\emph{G}}\\
		\hline
		Utility (\$) & 5 & 3 & 2 & 1& 4 & 2 & 1\\
		\hline
	\end{tabular}
	\label{table:utable}
	
\end{table}

Suppose there is a finite set of distinct items $I$ = \{$I_1$, $I_2$, $\cdots$, $I_m\}$, which is stored in a \textit{transaction database} $D$ = \{$T_1$, $T_2$, $\cdots$, $T_n$\}. Each transaction contains a subset of $I$ and a unique identified number, which can be abbreviated as $tid$. Each item in a transaction has a positive value $q(I_i,T_i)$, called its internal utility, which can also be interpreted as the occurrence of an item in the transaction. A set of definitions and properties is given as the follows:

\begin{definition}($U(X)$, utility of an itemset)
	\rm The external utility of an item is associated with the utility table, and can be considered as $v(I_i)$. And the external utility of an itemset $X$ is simply computed by $\sum_{I_i \in X}v(I_i)$. $u(I_i,T_i)$ measures the total utility of an item in a transaction, which is calculated by $q(I_i,T_i) \times v(I_i)$. The utility of an itemset $X$ in a transaction is calculated as ($min(q(x_j,T_i))$ $\times$ $v(X)$ where $x_1$, $\cdots$, $x_j$, $\cdots$, $x_n$ $\in$ $X$. Thus, the utility of an itemset is the sum of utility of this itemset in all transactions, i.e. $\sum_{x_j \in T_i}$($min(q(x_j,T_i))$ $\times$ $(\sum_{x_j \in X}v(x_j)$)). 
\end{definition}

For instance, in Table \ref{table:db}, the \textit{external utility} of $A$ is \$5, and the utility of $A$ in transaction $T_1$ is denoted by $u(A,T_1)$ = \$5 $\times$ 2 = \$10. Therefore, the utility of itemset $\{A,B\}$ in $T_1$ is (\$5 + \$6) $\times$ 1 = \$11, and the utility of itemset $AB$ is computed as $U(AB)$ = $U(AB, T_1)$ + $U(AB, T_2)$ = (\$5 + \$3) $\times$ 1 + (\$5 + \$3) $\times$ 2 = \$24. The transaction-weighted utility of itemset $X$ is denoted as $TWU(X)$ \cite{tseng2015efficient}. It is the sum of transaction utility of all the transactions containing $X$, which is defined as $TWU(X)$ = $\sum_{X \epsilon T_i, T_i \in D}$$TU(T_i)$, in which $TU(T_i)$ = $\sum_{x\in T_i \wedge X \subseteq T_i}$$u(x,T_i)$.
 
\begin{definition}($S(X)$, support of an itemset)
	\rm The support  (aka frequency) of an itemset $X$ is associated with the internal utility of an item, which is defined as $S(X)$ = $\sum_{X \subseteq T_i}q(X,T_i)$ = $\sum_{X \subseteq T_i}$$min(x_j, T_i)$ where $x_1$, $\cdots$, $x_j$, $\cdots$, $x_n$ $\in$ $X$.
\end{definition}

For instance, in Table \ref{table:db}, the TWU of item $A$ is computed by $TWU(A)$ = $TU(T_1)$ + $TU(T_2)$ = \$13 + \$23 = \$36. The frequency of itemset $C$ is calculated by $q(C)$ = $q(C, T_1)$ + $q(C, T_4)$ + $q(C, T_5)$ = 1 + 2 + 2 = 5.

\begin{definition} (High-utility itemset and high-frequency itemset)
	\rm An itemset is called a high-utility itemset (HUI) if its utility in a database is higher than or equal to a specified utility threshold, denoted as \textit{min\_util}. Similarly, an itemset is called a high-frequency itemset (HFI) if its support is higher than or equal to a specified frequency threshold, denoted as \textit{min\_fre}.
\end{definition}

Table \ref{table:db} shows a transaction database containing 5 transactions, where each letter represents a specific item, and the corresponding number represents the quantity in the corresponding transaction. The utility table is shown in Table \ref{table:utable}. If \textit{min\_util} is \$15, and \textit{min\_fre} is 3. Then a complete set of HUIs should be: \{$A$\}: \$15, \{$B$\}: \$24, \{$E$\}: \$16, \{$AB$\}: \$24, \{$BF$\}: \$15, \{$BDE$\}: \$16. And a complete set of HFIs is \{$A$\}: 3, \{$B$\}: 8, \{$C$\}: 5, \{$D$\}: 3, \{$E$\}: 4, \{$F$\}: 3, \{$G$\}: 4, \{$AB$\}: 3, \{$BF$\} :4, \{$FG$\}: 3.

\subsection{Problem statement}

With the significant progress in data mining and knowledge discovery, tremendous data mining algorithms have been applied in many fields such as business. With appropriate applications in marketing, salesmen can better develop their own strategies to obtain more profit.

Our study mainly focuses on pattern classification by applying the indicators of frequency and utility. Both of them are very essential in developing strategies from a transaction database because they are the fundamental information of an item or itemset. However, it is widely seen that utility or frequency is used as the sole measure in many typical applications, such as shopping basket analysis. Therefore, the following situations can be seen:  What if an item (or itemset) of high frequency has a quite low utility (low profit in business) in a transaction database? What if an item (or itemset) of high utility only occurs for a few times in a transaction database?

Both situations mentioned above indicates two types of itemsets, high-frequency itemsets utility and low-frequency itemsets with high utility. At the same time, there are also itemsets which can bring high profit and have high occurrences. Naturally, in most cases people are preferable to high-frequency itemsets with high utility. However, the first two types of itemsets can also have positive impact, with less importance than high-frequency itemsets with high utility. Therefore, these three types of itemsets actually play different important roles in strategies developing and rule designing.

To better discover the potential of all patterns, an acceptable solution is to classify patterns according to these two indicators simultaneously. Given the threshold of frequency and utility, there are four types of patterns \cite{wang2007pushing}. 

\begin{figure*}[h]
	\centering
	\includegraphics[height=0.2\textheight, width=0.4\textwidth]{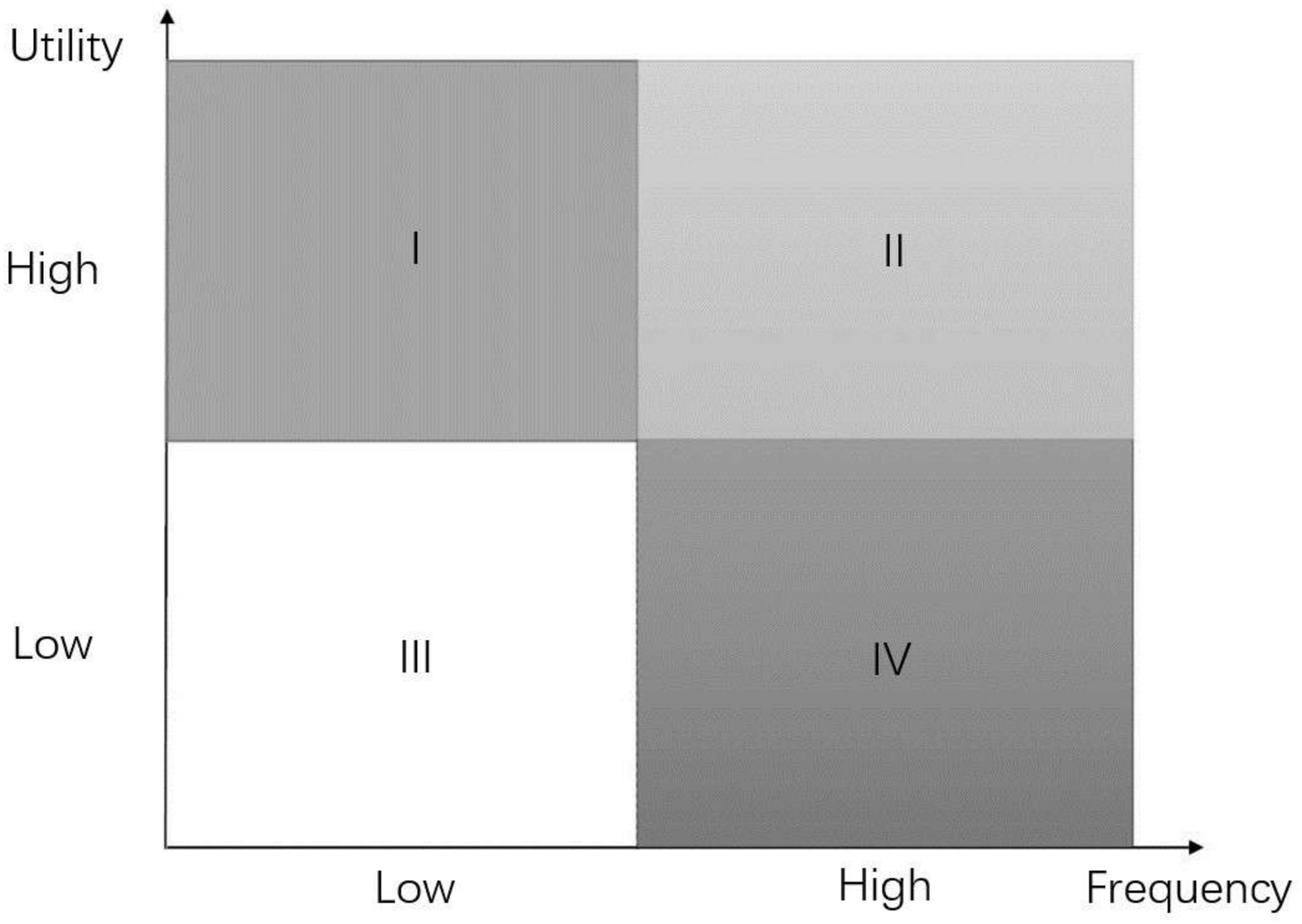}
	\caption{Four types of patterns \cite{wang2007pushing}.}
	\label{fig:four pattern}
\end{figure*}

\begin{definition}
	\rm By taking frequency and utility together into account, there are four types of patterns: High Frequency and High Utility itemset (\textit{HFHUI}); High Frequency and Low Utility itemset (\textit{HFLUI}); Low Frequency and High Utility itemset (\textit{LFHUI}); and Low Frequency and Low Utility itemset (\textit{LFLUI}). These four patterns can be illustrated in Fig. \ref{fig:four pattern}.
\end{definition}

However, among these four categories, LFLUI actually have little contribution to rule discovery and strategies designing. Therefore, the goal of our proposed algorithms is to collect the other three types of itemsets from the given database. Naturally, if the utility and frequency of an itemset can satisfy both thresholds then it belongs to HFHUI. If the itemset utility satisfies the threshold but the frequency does not, then it is HFLUI. Similarly, LFHUI and LFLUI are sorted in the same way. Recalled that in \textit{Example I}, we denote the minimum frequency threshold is 3 and the minimum utility threshold is \$15. In such case, we can obtain: \textit{HFHUI}: \{$A$\}: \$15, 3 (the first one represents utility, the second one represents frequency), \{$B$\}: \$24, 4, \{$E$\}: \$16, 4, \{$AB$\}: \$24, 3, \{$BF$\}: \$15, 4; \textit{HFLUI}: \{$C$\}: \$6, 3, \{$D$\}: \$3, 3, \{$F$\}: \$6, 3, \{$FG$\}: \$9, 3; \textit{LFHUI}: \{$BDE$\}: \$16, 2.

\begin{figure*}[h]
	\centering
	\includegraphics[scale=0.28]{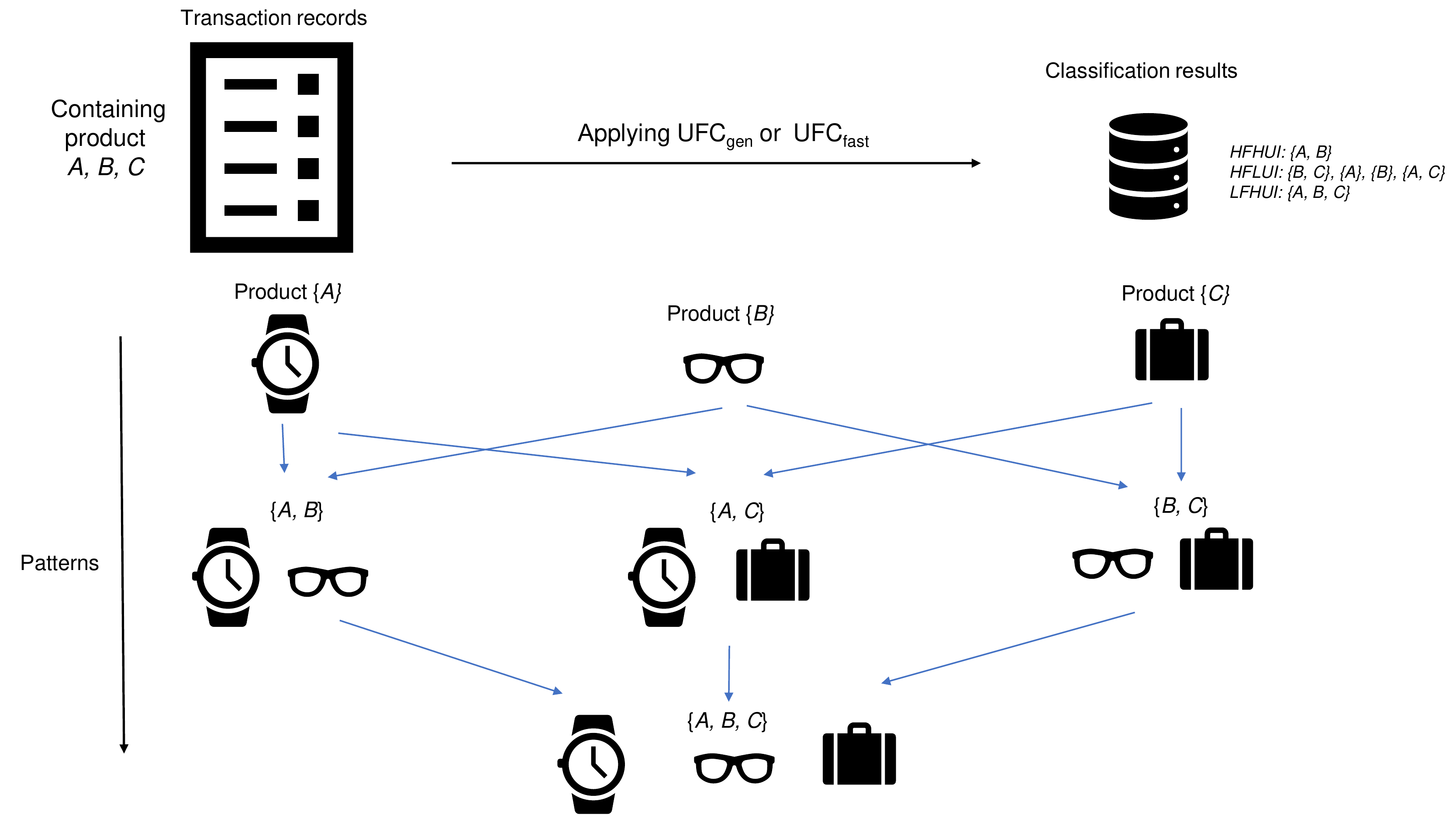}
	\caption{Classification processes in smart manufacturing.}
	\label{fig:process}
\end{figure*}

Generally, the application of our proposed algorithm in smart manufacturing can be split into the following steps: (1) Read the transaction records from the database; (2) Extract the necessary information from the records, i.e. the id of each good and its frequency and utility in each transaction; (3) Apply UFC$_{gen}$ or UFC$_{fast}$ on these records; (4) Obtain the classification results of three types of patterns; (5) Finally, proceed advanced analysis and design new strategies according to the classification results. These processes are shown in Fig. \ref{fig:process}.

\section{UFC$_{gen}$ Algorithm}
\label{algorithm1}
 
In this section, we introduce an algorithm called UFC$_{gen}$. The key idea is that we collect three types of itemsets based on the their utility and frequency in two phases. Details of this are presented below.

In the first phase, by utilizing the transaction-weighted utility downward closure property \cite{liu2005two} and frequency downward closure property, the UFC$_{gen}$ algorithm can efficiently generate new candidates and put them into the set of final candidates. In the second phase, an extra scan of the whole transaction database is needed to calculate the real utility of each itemset in the final candidates set. 

\subsection{Phase I}
 
In the first scan of the database, we record the TWU and frequency of each 1-itemset, and then generate new candidates by applying an operation called \textit{connection operation} level by level until there are no new candidates generated. Now we introduce some significant properties and definitions which are applied in Phase I.

\begin{property}(TWU downward closure property \cite{liu2005two})
	\rm  In a given database $D$, for any itemset $X$ $\in$ $D$, if $TWU(X)$ is less than a user-specified \textit{min\_util}, then all supersets of $X$ cannot be an HUI. If $X$ $\subseteq$ $X'$, then \textit{min\_util} > $TWU(X)$ $\geq$ $TWU(X')$. 
\end{property}

For example, assume that \textit{min\_util} is \$20, then in Table \ref{table:db} we can obtain \textit{TWU} of the itemset $E$, which is exactly \$16. Because \$16 $<$ \$20, by Property 1, we can conclude that $E$ cannot be an HUI.

\begin{property}(Frequency downward closure property \cite{agrawal1994fast})
	\rm  In a given database $D$, for any itemset $X$, if $X$ is a low-frequency itemset, then for any superset $X'$ of $X$, $X'$ can not be a high-frequency itemset.
\end{property}

For example, suppose that \textit{min\_fre} is 4, then by Table \ref{table:db} and Table \ref{table:utable} we obtain the support of the itemset $AB$, which is computed by $q(AB, T_1)$ + $q(AB,T_2)$ = 1 + 2 = 3. Since 3 $<$ 4, then by Property 2, we can conclude that any superset of $AB$ such as $ABC$ cannot be a HFI.

\begin{property}
	\rm  In a given database, We denote any high transaction-weighted utility itemset in $D$ as HTWUI. Then with the same user-specified \textit{min\_util}, if an itemset is an HUI, then it must be a HTWUI.
\end{property}

\begin{proof}
	\rm Let $D$ be the database and $W_i$ be the collection of all itemsets in transaction $T_i$, for a given itemset $X$, we define $X'$ = $W \backslash X$, then for any itemset $X \subseteq $ \textit{HUIs}, we have
	$ min\_util \leq U(X) $ = $\sum_{X\epsilon T_i\bigwedge T_i\epsilon D} U(X,T_i)$ $\leq \sum_{X'\epsilon T_i\bigwedge T_i\epsilon D}(U(X,T_i)+U(X',T_i)) $ $ = \sum_{X'\epsilon W_i\bigwedge W_i\epsilon D}TU(W_i)$ = $TWU(X)$.	Therefore, $X$ is also a HTWUI.
\end{proof}

For example, we still assume that \textit{min\_util} is \$20, from Table \ref{table:db} and Table \ref{table:utable}, we obtain the utility of itemset $B$, which is computed by $u(B, T_1)$ + $u(B, T_2)$ + $u(B, T_3)$ + $u(B, T_5)$ = \$3 $\times$ (2 + 3 + 2 + 1) = \$24. Because \$24 $>$ \$20, then by Property 3, we can conclude that $B$ must be a HTWUI.

\begin{definition}(Operation of connection)
	\rm we define a set as $C_k$ if for every itemset $I\subseteq C_k$, $I$ is an itemset containing $k$ items. We denote an operation as $\oplus$. For each $I_1$, $I_2$ $ \subseteq C_k$, if the condition satisfied that for $i \epsilon$ 1, 2, $\cdots$, $k-1$, $I_1[i]$ = $I_2[i]$,  $I_1[k]$ $\neq$ $I_2[k]$, then $I_1$ $\oplus$ $I_2$ = $I_3$, where $I_3$ = $\{I_1[1]$, $I_1[2]$, $\cdots$, $I_1[k]$, $I_2[k]\}$.
\end{definition}

\begin{table}[h]
		\caption{Characteristics of candidate sets}
	\centering
	\setlength{\tabcolsep}{4mm}{\begin{tabular}{ccc}
			\hline
			\textbf{Itemset} & \textbf{\textit{TWU}}& \textbf{Frequency}\\
			\hline
			$A$ & \$36 & 3\\
			
			$B$ & \$52 & 8\\
			
			$C$ & \$35 & 5\\
			
			$D$ & \$26& 3\\
			
			$E$ & \$16 & 2\\
			
			$F$ & \$22 & 3\\
			
			$G$ & \$16 & 4\\
			\hline
	\end{tabular}}
	\label{table:trans}	
\end{table}

Through the operation of connection, we can generate new candidates level-by-level, and any itemset in level $k$ would be a $k$-itemset. From Table \ref{table:db}, we can obtain each item's TWU and frequency in Table \ref{table:trans}. Now suppose \textit{min\_util} and \textit{min\_fre} is \$30 and 4, respectively, then by scanning Table \ref{table:trans}, we have to omit item $D$, item $E$ and item $F$ since the either of their frequency or utility satisfy the thresholds. Hence they cannot engage in connection operation. Hence, in the next level, we will only obtain itemset like $AB$, $AC$, $AG$ and so on connected by $A$, $B$, $C$ and $G$. In phase I, we continue this operation until no candidate sets can take part in the connection operation to enter the next level.


Using the properties and definitions above, we develop our pruning strategies to efficiently reduce search spaces. The strategies are presented the followings:

\begin{itemize}
	\item{\textbf{Strategy 1}}. If the TWU of current itemset is less than the utility threshold and the frequency of it is less than the frequency threshold, then current itemset cannot appear in phase II, which indicates the current itemset is automatically a LFLUI.
	
	\item{\textbf{Strategy 2}}. If the transaction-weighted utility of current itemset is higher than or equal to the utility threshold or the frequency of it is higher than or equal to the frequency threshold, we add the itemset to phase II. Then we apply the \textit{connection operation} to all such itemsets in order to generate new candidates.
\end{itemize}

In Phase I, the initial inputs are: \textit{$C_1$}, \textit{min\_util}, and \textit{min\_fre}. We use $cur$ to denote the sets of current itemsets with the proposition of TWU and frequency, denoted as $twu$ and $fre$, respectively.  When the length of $cur$ is zero, the algorithm automatically terminates. And the set $cand$ is used to represent the final candidate itemsets to be classified into three different types of itemsets. To be noticed, the null of a set means their is no itemset in the set. In lines 3-8, we apply our strategies in this algorithm. If the proposition of current itemset in $cur$ does not satisfy either thresholds, then we remove it from $cur$. In lines 10-19, we perform the \textit{connection operation} on all itemsets in $cur$. Based on these strategies, all the itemsets that satisfy the condition would be added to the final candidates $cand$, which will be classified in phase II. Details of the procedure in phase I are represented in Algorithm 1.

\begin{algorithm}[h]
	\caption{\textit{Phase I} of UFC$_{gen}$}
	\KwIn{
		$C_1$: the set of all 1-itemsets;   \textit{min\_util}, the threshold of utility;   \textit{min\_fre}, the threshold of frequency.
	}
	\KwOut{
		\textit{cand}: the set of final candidates.	
	}
	\BlankLine
	initialize $cur$ $\leftarrow$ $C_1$, \textit{cand} $\leftarrow$ null\;

	\While{\textit{\textit{length}}(cur) $>$ 0}{
		\For{each itemset $e$ in $cur$}{
			\If{\textit{e.twu} $\geq$ \textit{min\_util} or \textit{e.fre} $\geq$ \textit{min\_fre}}{
				add $e$ to \textit{cand}\;
				\Else{
					remove $e$ from $cur$\; 
				}
			}
		}
		$tmp$ $\leftarrow$ null\;
		
		\For{each itemset $I_1$ in \textit{cur}}{
			$k$ $\leftarrow$ length($I_1$)\;
			\For{each itemset $I_2$ after $L_1$ in $cur$}{
				\If{the first $(k-1)$ items in $I_1$ and $I_2$ are equal}{
					$I_3$ $\leftarrow$ $\{I_1[1]$, $I_1[2]$, ..., $I_1[k]$, $I_2[k]\}$\;
					add $I_3$ to \textit{cand}\;
					add $I_3$ to \textit{tmp}\;
				}
				
			}
		}
		clear \textit{cur}\;
		$cur$ $\leftarrow$ $tmp$\;
	}
	\Return{$cand$}
	\label{alo:phase 1}
\end{algorithm}

\subsection{Phase II}

An additional scan of the transaction database is required. The extra scan is designed to calculate the real utility of each candidate set. There are no pruning strategies in phase II. Its purpose is to collect three types of itemsets according the user-specified thresholds. Eventually, \textit{HFHUI}, \textit{HFLUI}, and \textit{LFHUI} are presented as the output. In Phase II, \textit{util} and \textit{fre} are denoted as the real utility and frequency of each itemset respectively in the set \textit{cand}. Details of phase II are given in Algorithm 2.

\begin{algorithm}[h]
\caption{\textit{Phase II} of UFC$_{gen}$}	

\KwIn{
	\textit{cand}: the final candidates after the \textit{Phase I};   \textit{min\_util}: the threshold of utility;   \textit{min\_fre}: the threshold of utility.
	
}   
\KwOut{
	\textit{HFHUI}: high frequency and high utility itemsets;   \textit{HFLUI}: high frequency and low utility itemsets;   \textit{LFHUI}: low frequency and high utility itemsets. 
}

initialize \textit{HFHUI} $\leftarrow$ null, \textit{HFLUI} $\leftarrow$ null, \textit{LFHUI} $\leftarrow$ null\;
scan the database again to obtain the real utility of each itemset\;
\For{each \textit{pattern} \textit{elem} in \textit{cand}}{
	\If{\textit{elem.util} $\geq$ \textit{min\_util} and \textit{elem.fre} $\geq$ \textit{min\_fre}}{
		add \textit{elem} to \textit{HFHUI}\;
	}
	\ElseIf {\textit{elem.util} $<$ \textit{min\_util} and \textit{elem.fre} $\geq$ \textit{min\_fre}}{
		add \textit{elem} to \textit{HFLUI}\;
	}
	\Else{
		add \textit{elem} to \textit{LFHUI}\;
	}
}
\Return{\textit{HFHUI}, \textit{HFLUI}, \textit{LFHUI}}
\label{alo:phase 2}
\end{algorithm}

\section{UFC$_{fast}$ Algorithm}
\label{algorithm2}

In this section, a more efficient algorithm called UFC$_{fast}$ is supplemented to solve the pattern classification problem. First we briefly introduce a special data structure called \textit{utility-list} \cite{liu2012mining}. The basic definitions are presented in the follows.

\subsection{Frequency-utility-list}

\begin{definition}(Revised transaction \cite{liu2012mining})
	\rm(1) A transaction is considered \textit{revised} if the transaction is sorted in ascending order according to the TWU; (2) all the itemsets whose TWU is less than the specified utility threshold is deleted from the transaction.
\end{definition}

\begin{definition}(Remaining utility \cite{liu2012mining})
	\rm In a database $D$, given a revised transaction and an item, the remaining utility of an itemset represents the sum of utility for items which are sorted after the current itemset. Let \textit{rutil}($X$) \cite{liu2012mining,lin2016fhn} denote the remaining utility of an itemset $X$ in a given database $D$, which is computed by $\sum_{X \subseteq T_i \wedge T_i \subseteq D}$$\sum_{x_j \succ X}u(x_j, T_i)$. Here $\succ$ means the position of $x_j$ is after $X$ in each revised transaction.
\end{definition}

For instance, by assuming the threshold of utility is \$30, we can rearrange Table \ref{table:db} according to the definition of revised transaction. The new table can be seen in Table \ref{table:rdb}. Considering the itemset $A$ in Table \ref{table:rdb}, the remaining utility of $C$ in $T_1$ can be computed as \textit{rutil}($C, T_1$) = $u(A, T_1)$ + $u(B, T_1)$ = \$5 $\times$ 1 + \$3 $\times$ 2  = \$11. For a list which stores i) the transaction number, ii) the utility of an itemset in a transaction, and iii) its remaining utility, we call such list as \textit{utility-list} \cite{liu2012mining}. This structure is very useful because it contains significant information, which helps avoid generating unpromising candidates. The utility-list \cite{liu2012mining} stores the transaction number, utility in a transaction and the remaining utility. Such structure is very useful because it contains very significant information and can be easily extended to a new itemset, and thus avoid generating new candidates.

\begin{table}[h]
	\centering
\caption{Example database (revised transaction)}
\begin{tabular}{ccc}
	\hline
	\textbf{tid}  & \textbf{Transaction} & \textbf{Utility}  \\
	\hline
	$T_1$ & ($C$, 1), ($A$, 1), ($B$, 2)&   \$13     \\
	$T_2$ & ($A$, 2), ($B$, 3)&   \$19        \\
	$T_3$ & ($B$, 2)&    \$6      \\
	$T_4$ & ($G$, 3), ($C$, 2)& \$7   \\
	$T_5$ & ($G$, 1), ($C$, 2), ($B$, 1)&  \$8 \\
	\hline
\end{tabular}
\label{table:rdb}
\end{table}

\begin{figure}[h]
	\centering
	\includegraphics[scale=0.25]{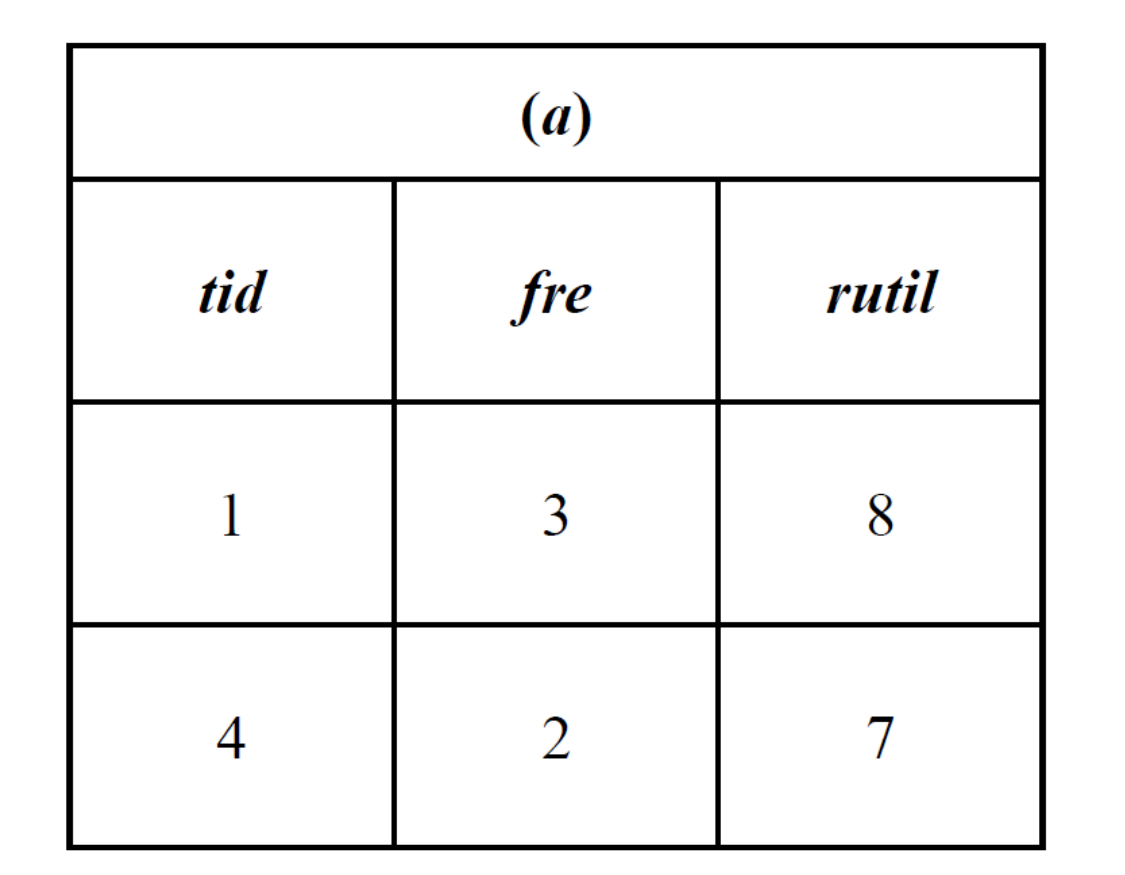}
	\caption{\textit{FU-list} of itemset $a$}
	\label{FU-list:a}
	
\end{figure}

To solve the pattern classification problem, we slightly change the utility-list, by substituting the second column into internal frequency of the item in each transaction, that is, the frequency of the item is recorded at each \textit{tid}. For the above example, suppose the external utility of item $a$ is \$2, then the list of a should be transformed into the list illustrated in Fig. \ref{FU-list:a}. We present it as a new data structure, called frequency-utility-list, or simply denotes it as FU-list.  

\begin{definition}(Frequency-utility list (FU-list))
	\rm For a given database, the FU-list contains a set of tuples, and each tuple has three fields: \textit{tid}, \textit{fre}, \textit{rutil}. The \textit{fre} is the occurrence of an itemset in the current transaction.
\end{definition}

To obtain the set of FU-lists, we have to change the definition of a revised transaction, now instead of deleting those itemsets with low TWU, we remove itemsets with low TWU and low frequency. Constructing the FU-lists for one-itemsets requires twice scan of the database. In the first scan, we calculate each one-itemset's TWU and frequency in order to transform each transaction into a revised transaction. Then in the second scan, we construct FU-lists from these new transactions.

\subsection{Frequency-utility-list of multiple itemsets}
 
Building the FU-list of an $k$-itemset is similar to the operation of connection mentioned in the UFC$_{gen}$ algorithm, where the intersection of transactions is not null and the first $k$-1 terms in two different $k$-itemsets are equivalent. We now introduce a new operation called \textit{1-extension}, which is used to construct a new FU-list of ($k$+1)-itemset from two different $k$-itemsets with the same prefix.

\begin{definition}(1-extension \cite{liu2012mining})
	\rm For two different $k$-itemsets having the same first ($k$-1)-items, denote them as \textit{ex} and \textit{ey}, where $e$ denotes the first ($k$-1)-items. Assume in the revised transaction the item $y$ is after item $x$, then by traversing their FU-list, if a shared transaction is found, we create a new entry by adding the information of shared \textit{tid}, the smaller frequency of \textit{ex} and \textit{ey} in this transaction, and the remaining utility of \textit{ey}. Then we add the entry to the FU-list of \textit{exy}. We keep the 1-extension operation until no more shared transactions are found. 
\end{definition} 

\begin{figure}[h]	
	\centering
	\includegraphics[scale=0.45]{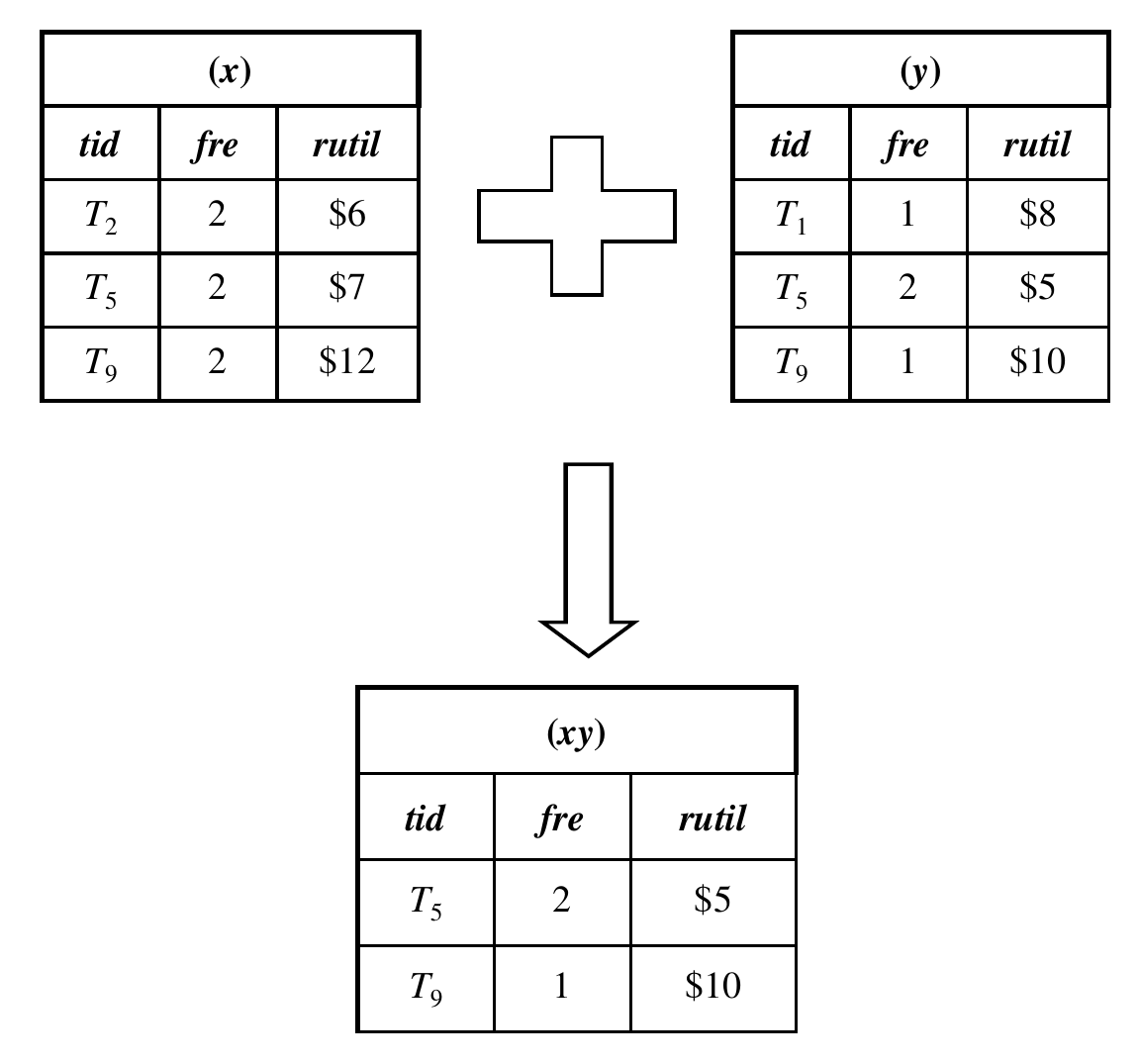}
	\caption{\textit{FU-list} of itemset \{$xy$\}}
	\label{FU-list:xy}
\end{figure}

Suppose there are two FU-lists created for two new items $x$ and $y$, whose TWU values are \$50 and \$100, respectively. It can be found that their common \textit{tids} are $\{5, 9\}$.  For the frequency of itemset $\{xy\}$ in $T_5$ and $T_9$, we choose the smaller frequency of item $x$ and item $y$. Therefore, in $T_5$, the frequency of $\{xy\}$ should be 1 and in $T_9$ it should be 2. As for the remaining utility of $\{xy\}$, because the TWU of $y$ is larger than $x$, then in the revised transaction $y$ should be after $b$, thus we have \textit{rutil}($xy$, $T_5$) = \textit{rutil}($y$, $T_5$) = \$5, and \textit{rutil}($xy$, $T_9$) = \textit{rutil}($y$, $T_9$) = \$10. Therefore, the \textit{FU-list} of $\{xy\}$ is built. The process is shown in Fig. \ref{FU-list:xy}.

\begin{property}(Remaining utility downward closure property)
	\rm  The property is similar with \textbf{Property 1}. If the sum of an itemset's utility and the remaining utility of the itemset is less than the utility threshold, then any 1-extension of the itemset $X$ cannot be an HUI, i.e., if $X \subseteq X'$ and \textit{U(X)} + \textit{rutil(X)} $<$ \textit{min\_util}, then $X'$ cannot be an HUI.	
\end{property}

\begin{proof}	 
	\rm In the database $D$, for an itemset $X$, let $X'$ be a 1-extension of $X$, define $D_{X'}$ as the set of revised transactions containing $X'$, and $D_{X}$ as the set of revised transactions containing $X$. Since $X'$ is a 1-extension of $X$, then $D_{X'}$ $\subseteq$ $D_{X}$. Furthermore, every item after itemset $X'$ must be included in the set of items after itemset $X$. Therefore, we can obtain: $ min\_util$ $>$ $U(X)$ + $\textit{rutil}(X)$  = $\sum_{T_i \in D_X}(u(X, T_i) + rutil(X, T_i))$ $\geq$ 
	$\sum_{T_j \in D_{X'}}(u(X, T_j) + \textit{rutil}(X, T_j))$ $\geq$ $\sum_{T_j \in D_{X'}}{u(X', T_j)}$ = $U(X')$.
	
\end{proof}

\begin{algorithm}[h]
	 
\caption{\textit{Extend} procedure}
\KwIn{
	$e$: an item;   \textit{ex.FU}: the \textit{FU-list} of itemset $ex$;   \textit{ey.FU}: the \textit{FU-list} of itemset $ey$.
}
\KwOut{
	\textit{exy.FU}: the \textit{FU-list} of itemset \textit{exy}.
}

initialize \textit{exy.FU} $\leftarrow$ null\;
\For{each entry $a$ $\epsilon$ \textit{ex.FU}}{
	\If{$\exists$ \textit{a.tid} == \textit{b.tid} $\wedge$ $b$ $\in$ \textit{ey.FU}}{
		\textit{fre} $\leftarrow$ \textit{min(a.fre, b.fre)}\;
		create a new entry \textit{elem} $\leftarrow$ \textit{[a.tid, fre, b.rutil]}\;
		\textit{exy.FU} $\leftarrow$ \textit{exy.FU} $\bigcup$ \textit{elem}\;
	}
}
\Return{\textit{exy.FU}}	
\label{alo:extend}

\end{algorithm}

In the extended procedure, we take $e$, \textit{ex.FU}, and \textit{ey.FU} as input, where $e$ is an item (can be empty), and $y$ is the item after the item $x$ in the revised transactions. The remaining utility of an itemset in each transaction is denoted by \textit{rutil}. In lines 2-3, we collect transactions where $ex$ and $ey$ both exist. In lines 3-5, for the construction of new FU-list \textit{exy.FU}, for each satisfying transaction, we create a new entry for them. Then we select the smaller frequency of $ex$ and $ey$ as the frequency in the entry. Furthermore, the remaining utility of $ey$ is chosen as the new remaining utility of \textit{exy}. Finally, we append the new entry to the FU-list of \textit{exy}. Details of the extended procedure for the 1-extension of $ex$ are presented in Algorithm 3.

\subsection{Pruning strategies in UFC$_{fast}$}

The pruning strategies are slightly different from the strategies in UFC$_{gen}$ because of the special structure of the FU-list. There is an extra remaining utility, which can be used to derive a new but similar pruning strategy. In the FU-list, the frequency is already denoted in the second column; thus, the itemset utility can be easily obtained by acquiring the external utility of the itemset through the corresponding utility-table, and thus we can immediately identify its category. Now we introduce a lemma, which is used as one of the pruning strategies in the UFC$_{fast}$ algorithm.

\begin{lemma}
	\rm  Given the FU-list of an itemset $X$ with its external utility, if the product of its frequency and external utility (the utility of the itemset) plus its remaining utility is less than the given minimum threshold, then any extension $X'$ of $X$ is not an HUI.
\end{lemma}

Taking the \textit{FU-list} of $x$ as an instance, suppose its external utility is \$1 and the utility threshold is \$35. And its frequency multiples external utility plus its remaining utility is equal to $u(x,T_2)$ + $(x,T_5)$ + $(x,T_9)$ = (2 $\times $ \$1 + \$6) + (2 $\times$ \$1 + \$7) + (2 $\times$ \$1 + \$12) = \$31 $<$ \$35. Then any extension of $x$ can not be a HUI. With \textbf{Lemma 1} and \textbf{Property 1} and \textbf{Property 2} mentioned in UFC$_{gen}$ algorithm, the pruning strategies are developed as the follows:

\textbf{Strategy 3.} In the first scan of the database, we select those 1-itemsets whose transaction-weighted utility are higher than or equal to the utility threshold or those whose frequency is higher than or equal to the frequency threshold. If neither of the thresholds are achieved, then the 1-itemset and its extension must be a LFLUI, and thus we simply remove it from the current transaction.

For example, suppose the \textit{min\_util} is \$25 and \textit{min\_fre} is 5. According to Table \ref{table:db} and Table \ref{table:utable}, we obtain the TWU of itemset $F$ and frequency of it are computed by $U(T_4)$ + $U(T_5)$ = \$10 + \$12 = \$22 and $q(F, T_4)$ + $q(F, T_5)$ = 1 + 3 = 4 respectively. Because \$22 < \$25 and 4 < 5, then itemset $F$ would be a LFLUI, and we should delete it.
 	
\textbf{Strategy 4.} For each itemset, we can immediately obtain its real frequency and utility through its FU-list and utility table in the database, and thus the itemset can be classified immediately.   
 
For example, as shown in Fig. \ref{FU-list:a}, the FU-list of item $a$, by assuming the external utility of $a$ is \$2, we can obtain the frequency and utility of $a$: 5, \$10. If \textit{min\_fre} and \textit{min\_util} are 4 and \$15 respectively, then $a$ should be a HFLUI.

\textbf{Strategy 5.} If the current itemset's real frequency is higher than or equal to the frequency threshold or the sum of its utility plus and remaining utility is higher than the utility threshold, then we can apply 1-extension to the current itemset with other itemsets after it.   

For example, we assume that \textit{min\_util} is \$30 and \textit{min\_fre} is 3. According to Table \ref{table:utable} and Table \ref{table:rdb}, we obtain the sum of itemset $G$'s utility and its remaining utility is computed by $U(G)$ + \textit{rutil}($G$) = $(u(G, T_4)$ + $u(G, T_5)) $ + (\textit{rutil}($G, T_4$) + $\textit{rutil}(G, T_5))$ = \$1 $\times$ (3 + 1) + (\$2 $\times$ 2 + \$2 $\times$ 2 + \$3 $\times$ 1) = \$15. The support of $G$ is computed by $q(F, T_4)$ + $q(F, T_5)$ = 3 + 1 = 4. Because \$15 $<$ \$30 and 4 $>$ 3, we should create an 1-extension of $G$, including itemset $GC$, $GA$, and $GB$.

\subsection{Proposed UFC$_{fast}$ algorithm}
 
After twice scans of the database, we build the initial FU-list for all the one-itemsets. The operation of 1-extension starts from the set of original FU-lists. At each time, we first obtain the current itemset's frequency and utility and sort it according to the given thresholds. Then, by applying \textbf{Strategy 5}, we extend the current itemset with other itemsets. The UFC$_{fast}$ algorithm terminates when there are no more candidates can be generated. Details of the algorithm are presented in Algorithm 4.

\begin{algorithm}
\caption{ UFC$_{fast}$ algorithm}

\KwIn{
	$e$: an item, initially empty;   $FUs$: the set of \textit{FU-lists} of itemset $e$'s 1-extensions;   \textit{min\_util}: the threshold of utility;   \textit{min\_fre}: the threshold of frequency.
}
\KwOut{
	\textit{HFHUI}: high frequency and high utility itemsets;   \textit{HFLUI}: high frequency and low utility itemsets;   \textit{LFHUI}: low frequency and high utility itemsets.
}

initialize \textit{HFHUI} $\leftarrow$ null, \textit{HFLUI} $\leftarrow$ null, \textit{LFHUI} $\leftarrow$ null\;
\For{each \textit{FU-list} $t_1$ in $FUs$ }{
	x $\leftarrow$ the itemset corresponding to $t_1$\;
	classify itemset $x$\;
	\If{\textit{x.fre} $\times$ \textit{x.util} + \textit{x.rutil} $\geq$ \textit{min\_util} or $x.fre$ $\geq$ \textit{min\_fre}}{
		\textit{exFUs} $\leftarrow$ null\;
		\For{\textit{FU-list} $t_2$ after $t_1$ in $FUs$}{
			$y$ $\leftarrow$ the itemset corresponding to $t_2$\;
			\textit{exFUs} $\leftarrow$ \textit{exFUs} $\bigcup$ \textit{Extend}($e$, $x$, $y$)\;
		}
		call UFC$_{fast}$($x$, \textit{exFUs}, \textit{min\_util}, \textit{min\_fre})\;
	}
}
\label{alo:fast}
\end{algorithm}
 
In this algorithm, an item $e$ and the set of FU-lists \textit{FUs} are used as the input. At first, we scan every FU-list of $t_1$ in \textit{FUs}, the FU-lists of 1-itemsets are constructed for those whose transaction-weighted utility or frequency satisfy the minimum threshold, thus we can immediately sort the current itemset $x$ to corresponding category through its FU-list, this is exactly the process of \textbf{Strategy 3} and \textbf{Strategy 4}. On the other hand, in lines 5-10, an empty set of FU-lists \textit{exFUs} is created, and the \textbf{Strategy 5} is applied. If the sum of utility and remaining utility for $x$ or its frequency meets the minimum value, then we apply the \textit{extended} algorithm to it and all the itemset $y$ after it in order to create new FU-lists. Once a new FU-list is created, it is added to the \textit{exFUs}. Finally, the algorithm is called recursively, and it terminates when there is no more itemset that can be classified.

According to \textbf{Property 2} and \textbf{Property 4}, the \textbf{completeness and correctness} of the proposed UFC$_{fast}$ algorithm is satisfied; hence, it can ensure that each itemset is sorted into the corresponding category. Furthermore, all the itemsets can be applied to the 1-extensions operation if the condition is satisfied, hence the algorithm can guarantee that all the possible itemsets would be discovered.

\section{Experiments}
\label{experiments}
 
We performed extensive experiments to evaluate the UFC$_{gen}$ and UFC$_{fast}$ algorithms. Because the model is the first model that collects three types of itemsets according to frequency and utility, we can only compare these two algorithms on different datasets, in terms of execution time, memory consumption and scalability. 

Both algorithms are implemented in Java language. All experiments were conducted in a personal computer with 4 GB of RAM, running the 64-bit Microsoft Windows 10 operating system. In the experiments, the execution time, memory consumption, classification results and scalability analysis are evaluated respectively. In order to visualize the impact of thresholds on other factors, \textit{min\_util} and \textit{min\_fre} are represented as percentage in all the experiments. For instance, if the sum of utility in a database is \$1,000 and the minimum utility threshold is \$150, then \textit{min\_util} should be 15\%, which is calculated as $\frac{\$150}{\$1000}\times 100\%$.

In the experiment, unlike the general experiments for HFIM and HUPM, the parameters of thresholds are set in the form of combinations, which means that both frequency and utility are taken into account in the same experiment. In each experiment, we choose different thresholds of frequency and vary the threshold of utility. For each dataset, the parameters of both thresholds are all set at a very low level (less than 1\%), because in earlier testings we found that when the thresholds were set at a relative high level, then many itemsets could not meet the condition, which could cause difficulties in comparison for the proposed algorithms. From our previous testings, it is proved that an increase of 5\% for each point is a relative appropriate setting to show the general trend for the proposed algorithms.

\subsection{Experiment datasets}

Both synthetic and real datasets (foodmart, retail, T40I10D100K, and yoochooseBuys) were used in the experiments. The resource link for the synthetic datasets and real dataset are http://www.philippe-fournier-viger.com/spmf and recsys.acm.org/recsys15/challenge/. For the dataset foodmart, it has 21,556 rows of transactions, while having around 1,559 items contained. For datasets retail and T40I10D100K, the quantities of their transactions are much larger. At the same time, there are only averagely 4 items and 10.3 items for most of the transactions from foodmart and retail, separately, but for dataset T40I10D100K, it includes more than 70 items in each row of transactions. For all datasets, both internal utility and external utility are assigned to each item. The detailed characteristics of all datasets are listed in Table \ref{tab:data}, where $\#|D|$ represents the number of transactions in the database, and $\#|I|$ indicates the number of items in the database.


\begin{table}[h]
	\centering
	\caption{Characteristics of test datasets}
	\setlength{\tabcolsep}{4.3mm}{
		\begin{tabular}{l c c c}
			\hline
			\textbf{Dataset} & \textbf{$\#|D|$} & \textbf{$\#|I|$} & \textbf{AvgLen}\\
			\hline
			retail	     & 88,162  & 16,470  & 10.3 \\ 		
			foodmart     & 21,556  & 1,559   & 4.0\\
			T40I10D100K  & 100,000 & 942     & 39.6\\ 
			yoochooseBuys & 507,746   & 19,102 & 2.3\\
			\hline
	\end{tabular}}
	\label{tab:data}
\end{table}

\begin{figure*}[h]
	
	\begin{minipage}[ht]{0.18\textwidth}
		\scalebox{1}[1.1]{\includegraphics[height=0.7\textheight,width=3.75cm, trim=90 90 0 0]{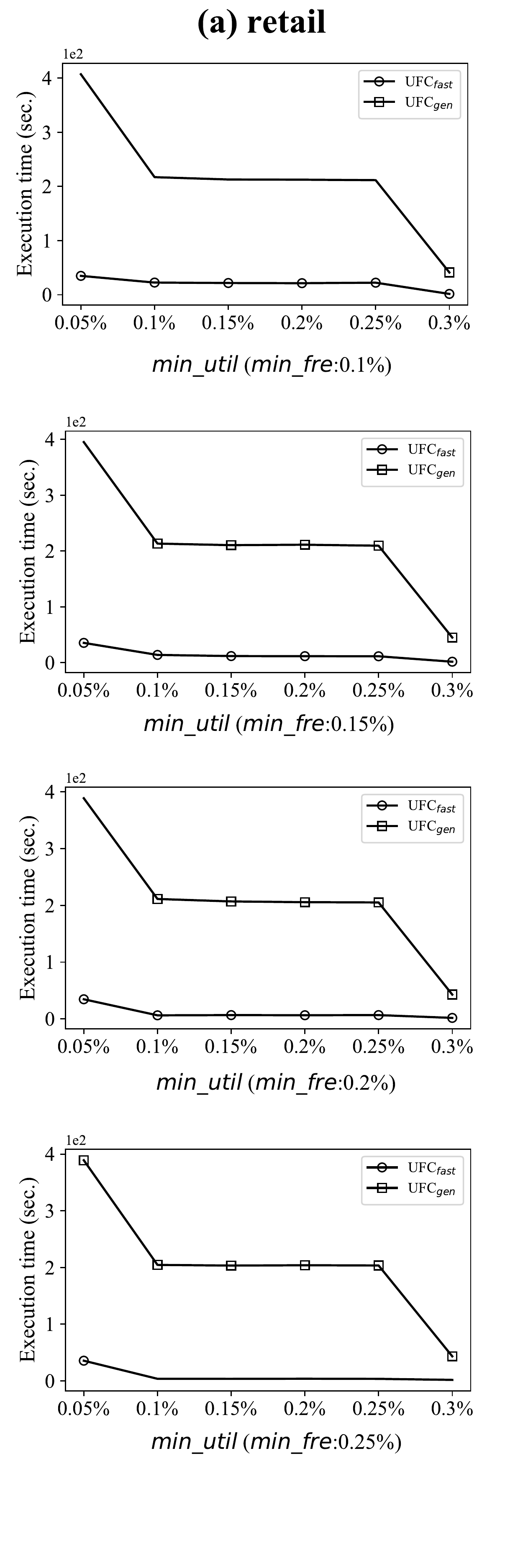}}
	\end{minipage}
	\hspace{18mm}
	\begin{minipage}[ht]{0.18\textwidth}
		\scalebox{1}[1.1]{\includegraphics[height=0.7\textheight,width=3.85cm, trim=60 90 0 0]{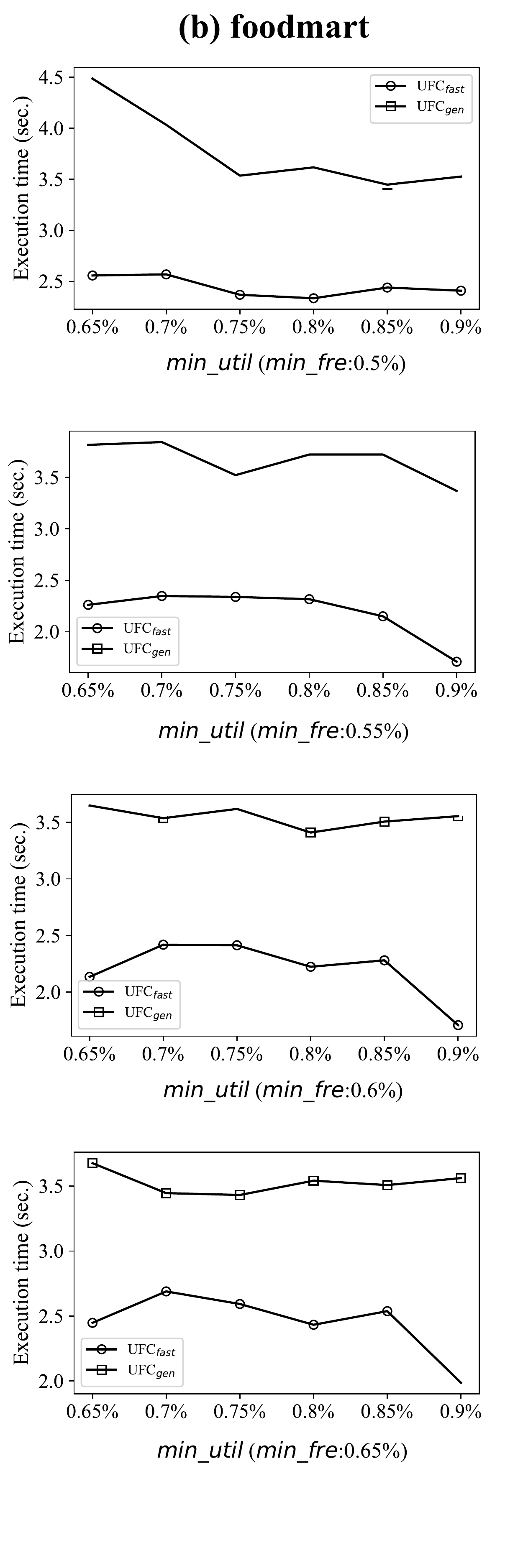}}
	\end{minipage}
	\hspace{18mm}
	\begin{minipage}[ht]{0.18\textwidth}
		\scalebox{1}[1.1]{\includegraphics[height=0.7\textheight,width=3.8cm, trim=60 90 0 0]{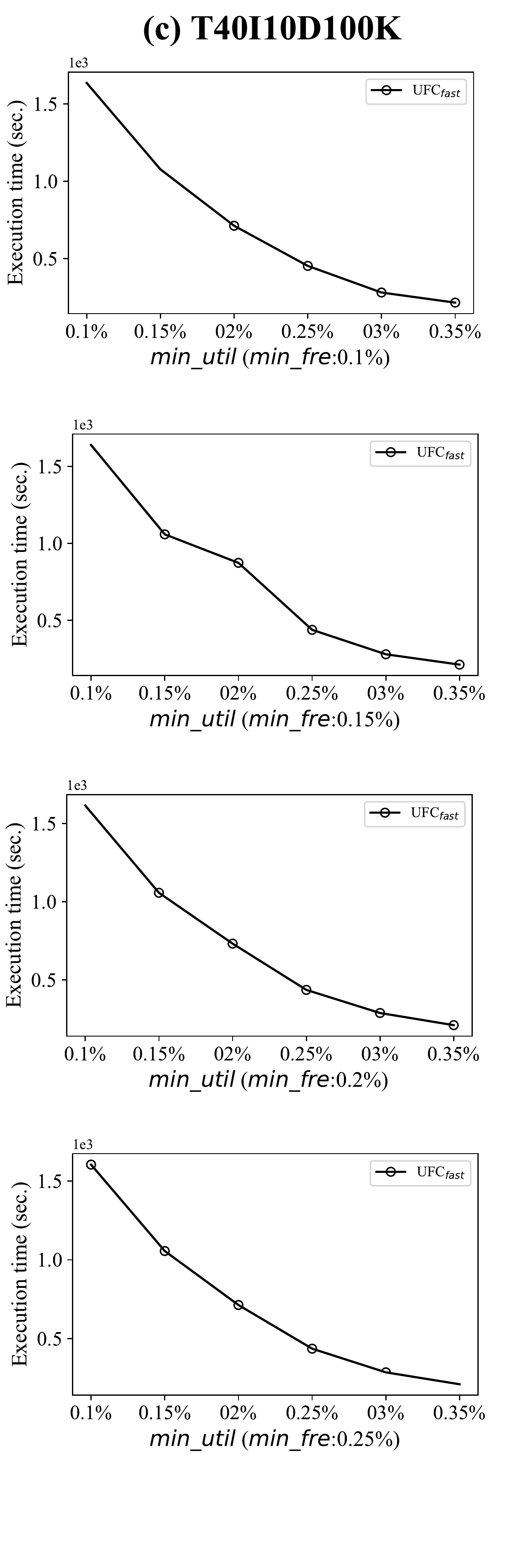}}
	\end{minipage}
	\caption{Execution time w.r.t different combination of thresholds of frequency and utility.}
	\label{fig:running time}
\end{figure*}

\subsection{Execution time}

The running time of UFC$_{gen}$ and UFC$_{fast}$ algorithms on synthetic datasets are illustrated in Fig. \ref{fig:running time}. A task would terminate if its running time exceeded 10,000 s. By setting a group of thresholds for both frequency and utility, we obtain the execution time of the proposed algorithms at each point. It is likely that the lower the threshold is, the more computation time is required for both algorithms. The execution time increases because there are more itemsets satisfying the requirements. For instance, on dataset retail, when \textit{min\_fre} and \textit{min\_util} are both 0.15\%, the running times for UFC$_{gen}$ and UFC$_{fast}$ are 210 s and 6.5 s, respectively. However, when the thresholds both become 0.1\%, their execution time became 217 s and 23 s, respectively.  

It is noticeable that UFC$_{fast}$ always requires less time than that on UFC$_{gen}$. This is mainly because UFC$_{gen}$ keeps generating new candidates level by level, during which a large amount of time is consumed in getting new candidates by applying the connection operation. In particular, when the number of items is extremely large, the computation time for combining current itemsets can be very slow. When the number of itemsets becomes smaller, the UFC$_{gen}$ algorithm can spend less time on connecting the old itemsets to generate new ones, thereby reducing the time difference between both algorithms. For example, from Fig. \ref{fig:running time}(b), it can be seen that when applying the dataset foodmart, the difference of running time between UFC$_{fast}$ and UFC$_{gen}$ is much smaller than that on dataset retail. The ranges for both proposed algorithms are within 5 seconds, and the largest time consumed is 2.6 s and 4.5 s, respectively, when the thresholds of frequency and utility are 0.1\% and 0.05\%, separately.

Fig. \ref{fig:running time}(c) shows the running time on dataset (T40I10D100K). Because the average length of the transaction database is large, for the UFC$_{gen}$, it has to deal with an enormous number of itemsets when setting the same thresholds as in dataset retail; in that case, the execution time is remarkably large, which means it has to be terminated. By contrast, for UFC$_{fast}$, it finishes the classification tasks longer than on dataset retail and foodmart, where the most time consumed is 1635 seconds when the thresholds are both 1\%. The overall trend of execution time for  UFC$_{fast}$ on it is relatively smooth. 

Overall, the proposed UFC$_{fast}$ has better performance than UFC$_{gen}$ on computation time under various thresholds of utility and frequency on each dataset. This is particularly evident when the database contains a large number of items and  many rows of transactions. On the other hand, from the figures it can be seen that the execution time of both algorithms does not always reduce linearly as the thresholds increase, but the fluctuations are often acceptable.

\subsection{Memory consumption}

\begin{figure*}[ht]
	\begin{minipage}[ht]{0.18\textwidth}
		\scalebox{1}[1.0]{\includegraphics[height=0.7\textheight,width=3.5cm, trim=90 50 0 0]{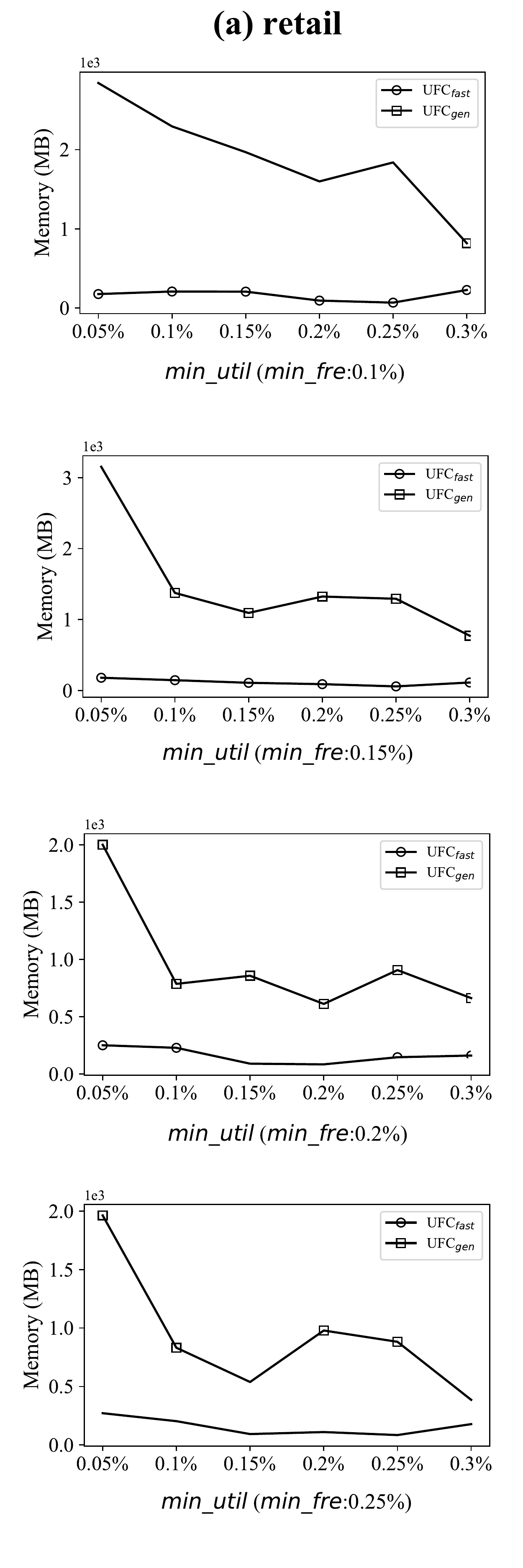}}
	\end{minipage}
	\hspace{18mm}
	\begin{minipage}[ht]{0.18\textwidth}
		\scalebox{1}[1.0]{\includegraphics[height=0.7\textheight,width=3.34cm, trim=90 50 20 0]{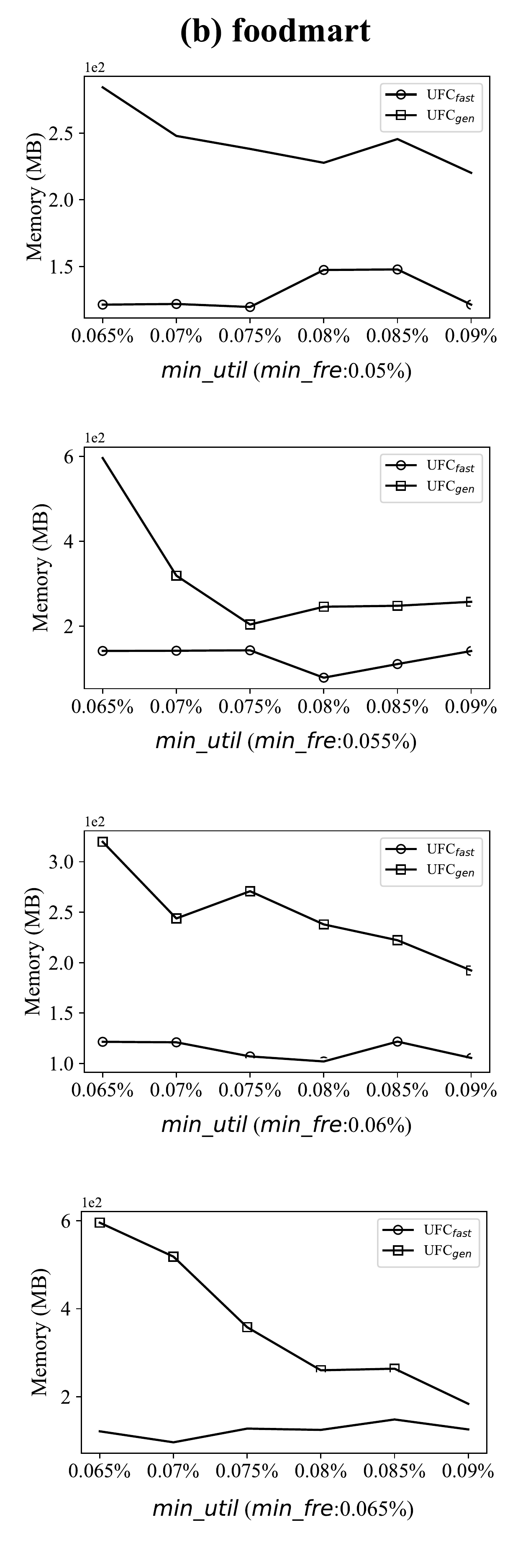}}
	\end{minipage}
	\hspace{18mm}
	\begin{minipage}[ht]{0.18\textwidth}
		\scalebox{1}[1.0]{\includegraphics[height=0.7\textheight,width=3.35cm, trim=80 50 20 0]{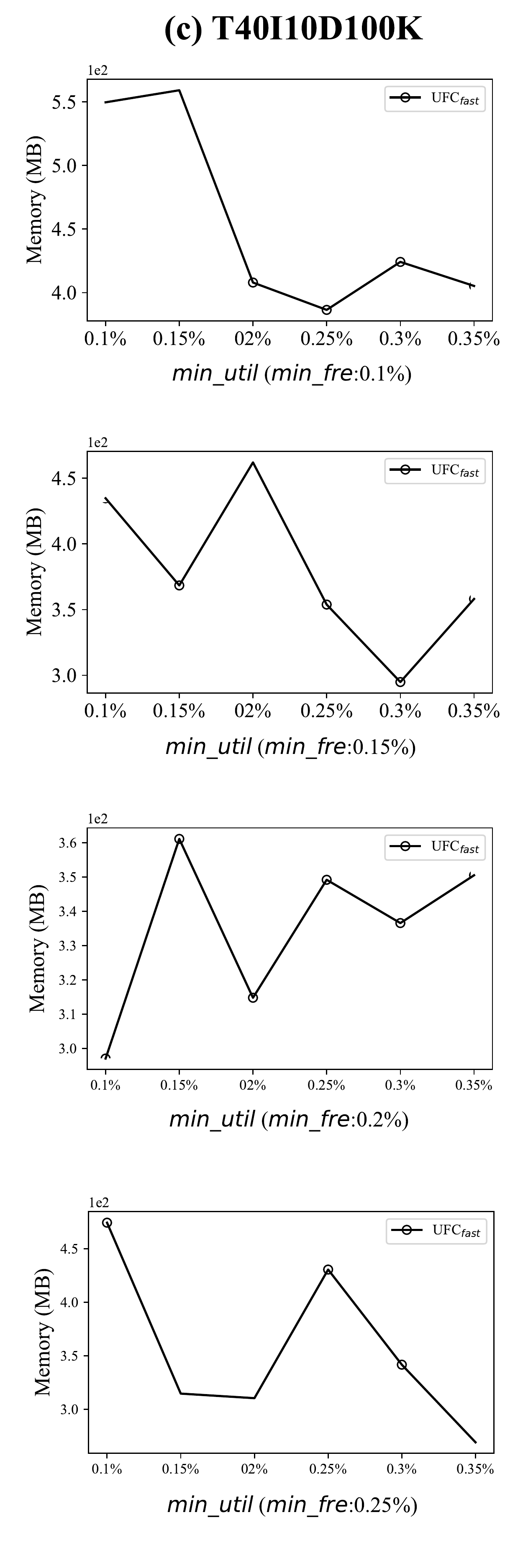}}
	\end{minipage}
	\caption{Memory consumption w.r.t different thresholds of frequency and utility.}
	\label{fig:memory}
\end{figure*}

Fig. \ref{fig:memory} shows the peak memory consumption of both algorithms on real and synthetic datasets. It can be seen that UFC$_{fast}$ always consumes much less memory than UFC$_{gen}$. This is mainly because UFC$_{fast}$ does not generate candidates, while UFC$_{gen}$ always generates new candidates from previous candidates, satisfying conditions each time.

Generally, the trends of memory consumption for both algorithms on the datasets are not linear. The peak memory of UFC$_{gen}$ fluctuates significantly on datasets retail and foodmart. When applying UFC$_{fast}$ to synthetic datasets, the memory consumption for it tends to be more stable. For instance, on dataset retail, with the increase of one of the thresholds, the memory consumption by applying UFC$_{fast}$ is maintained at approximately 200 MB, reaching the highest (271.78 MB) when the thresholds of frequency and utility are 0.3\% and 0.25\% respectively. By contrast, the figure for UFC$_{gen}$ experiences some significant fluctuations when situated in different thresholds. When the utility threshold is 0.05\%, its memory consumption of it is dramatically higher than that in other situations. While increasing the utility threshold, the memory consumed drops noticeably, but it does not maintain the declining trend, and increases in other situations. On the other hand, large fluctuations can be seen in UFC$_{fast}$ on T40I10D100K.

It can be concluded that the UFC$_{gen}$ algorithm may not fit in datasets containing a high quantity of items or transactions of a long average length. Because it means that there is likely to be a remarkable memory consumption on storage for new itemsets generated by connecting the current candidate sets. For instance, when we apply UFC$_{gen}$ on dataset T40I10D100K, the memory consumption is so large that exceeds the stack capacity of JVM, when the threshold of utility is 0.1\%. However, for the UFC$_{fast}$ algorithm, it stores information about all itemsets using the FU-list. At each time it directly classifies the current itemset using the pruning strategies without generating new candidates. Therefore, its memory consumption of it on different datasets always has a much better performance than that of UFC$_{gen}$. It only requires a little extra memory to store the remaining utility of each itemset. Overall, as in the experiment of testing execution time, UFC$_{fast}$ outperforms than UFC$_{gen}$ dramatically, and the main reason for this is that the former does not generate new candidates, while the latter does.

\subsection{Classification results}

By selecting different thresholds for frequency and utility, there are various classification results. Both algorithms obtained the same classification results for the datasets. Table \ref{tab:class} compares the number of \textit{HFHUI}, \textit{HFLUI} and \textit{LFHUI} for the tested datasets, with fixed frequency threshold and varying utility thresholds. The chosen frequency threshold from each dataset is denoted in the racket of the first column in the table. 

For dataset retail, the number of \textit{HFHUI} only decreases from 97 to 54 when the threshold increased, while the quantity of \textit{LFHUI} decreases dramatically from 293 to 24. Similarly, it can be seen that the number of \textit{LFHUI} on dataset T40I10D100K is 23,347 when \textit{min\_util} is 0.1\%, which is significantly large. However, the quantity drops rapidly to 3,086 when the threshold rises to 0.15\%. As for dataset \textit{foodmart}, it can be seen that the number of \textit{HFLUI} accounts for the majority of total candidates in all situations, which indicates that a large number of itemsets in the dataset tends to have the preposition of higher frequency and lower utility. With an increase in the threshold of utility, the numbers of \textit{HFLUI} and \textit{HFHUI} both decline, and the quantity of \textit{LFHUI} increases significantly.

\begin{table}[h]
	\centering
	\caption {Classification results on tested datasets}
	\setlength{\tabcolsep}{0.4mm}{\begin{tabular}{l c c c c}
			\hline
			\textbf{Dataset} & \textbf{\textit{min\_util}} & \textbf{$\#$ \textit{HFHUI}} & \textbf{$\#$ \textit{HFLUI}}& \textbf{$\#$ \textit{LFHUI}} \\ \hline
			
			\multirow{4}{*}{Foodmart (0.07\%)}
			&0.070\% &224 & 200 & 400 \\ 	
			&0.075\% &211&213 &333\\ 
			&0.080\% &187& 237& 273\\ 
			&0.085\% &170 &254 &214\\ 
			\hline
			\multirow{4}{*}{Retail (0.1\%)}
			&0.10\% & 97 & 22 & 293 \\
			&0.15\% &79 &40 &107\\ 
			&0.20\% & 66& 53& 47\\
			&0.25\% &54 &65 &24\\ \hline
			\multirow{4}{*}{T10I4D100K (0.1\%)}
			&0.10\% & 243 & 150 & 23,347 \\
			&0.15\% &188 &205 &3,086\\ 
			&0.20\% & 138& 255& 361\\
			&0.25\% &105 &288 &59\\ \hline
	\end{tabular}}
	\label{tab:class}
\end{table}

\subsection{Scalability analysis}

\begin{figure*}[!htbp]
	\centering	
	\scalebox{1}[1]{\includegraphics[height=0.18\textheight,width=0.85\textwidth]{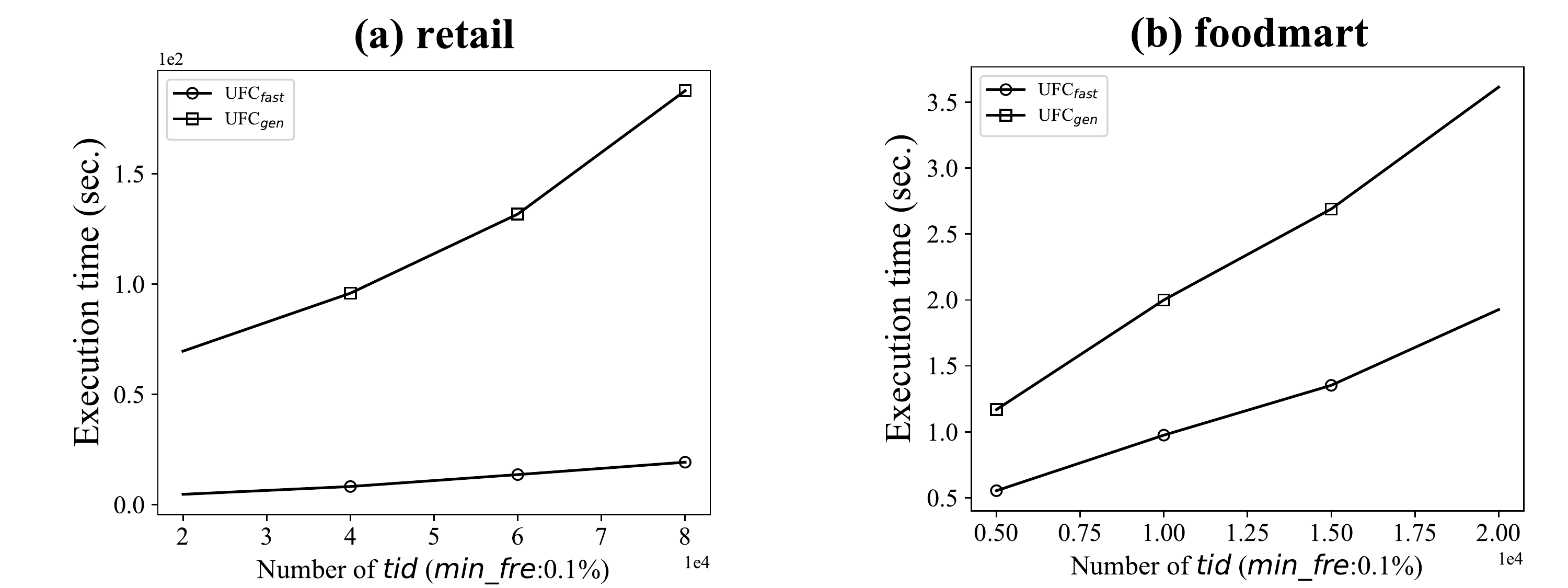}}
	\caption{Scalability w.r.t. execution time}
	\label{fig:scalar1}
\end{figure*}

\begin{figure*}[!htbp]	
	\centering	
	\scalebox{1}[1]{\includegraphics[height=0.18\textheight,width=0.85\textwidth]{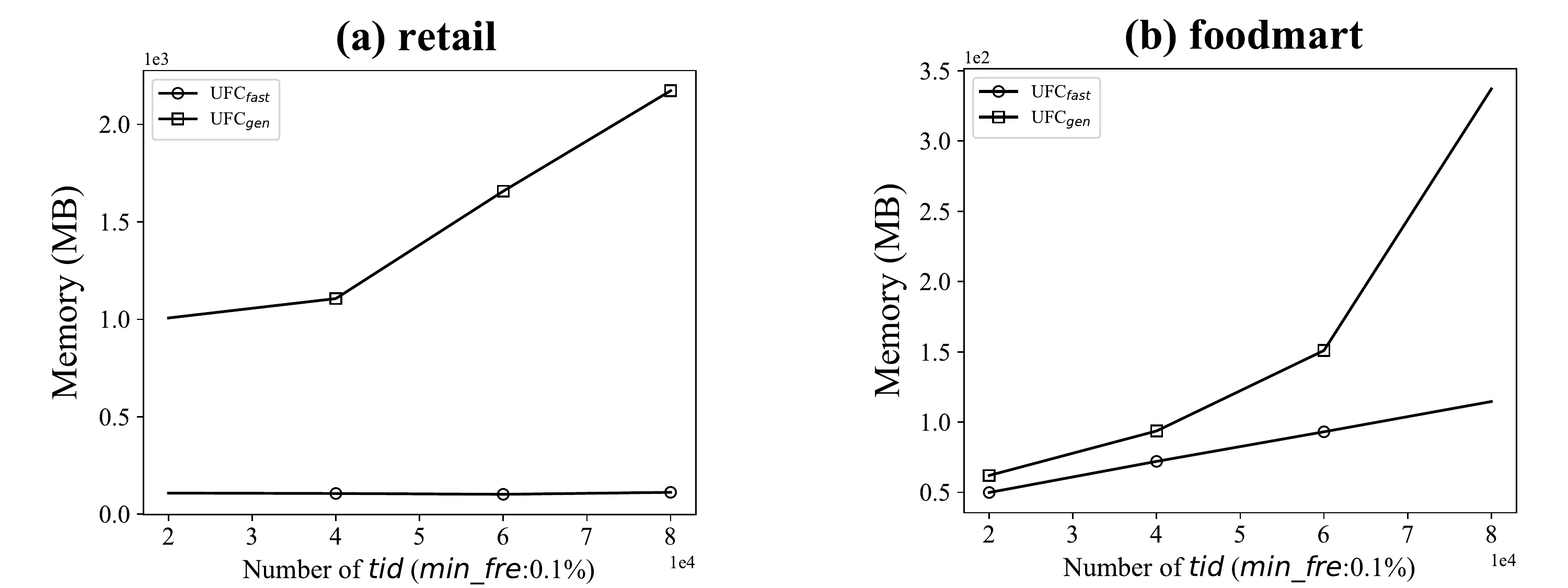}}
	\caption{Scalability w.r.t. memory consumption}
	\label{fig:scalar2}
\end{figure*}

We use two datasets (foodmart and retail) to perform the scalability analysis of the proposed algorithms. We fix the thresholds of utility and frequency to 0.1\%, while increasing the number of transactions each time. With the increase of in the number of transactions, the number of candidates could see a growing trend, because some candidates may not be included in those new additive transactions. Fig. \ref{fig:scalar1} and Fig. \ref{fig:scalar2} illustrate the execution time and memory consumption of UFC$_{fast}$ and UFC$_{gen}$ on dataset foodmart and retail.

Both proposed algorithms have excellent scalability with regard to execution time. On foodmart, we find that the execution time of both algorithms increases almost linearly, although the time for UFC$_{gen}$ is higher than UFC$_{fast}$ throughout the period. However, this gap is not noticeable. For dataset retail, the increment of transactions is 10000 each time. Changes in UFC$_{gen}$ become larger when applying different sizes of total transactions, spending 69.637 s when the size is 2,000, while it becomes 187.717 s when the size is 8000. On the other hand, the time spent by UFC$_{fast}$ seems to have a linear trend. The time grows smoothly and slowly, and in each situation, the time consumed is within 25 s, which is much less than the time spent by UFC$_{gen}$. 

On the other hand, in Fig. \ref{fig:scalar2}, it shows the memory consumption of UFC$_{fast}$ and UFC$_{gen}$ on both datasets, with the same division of transactions as before. For UFC$_{fast}$, it tends to keep unchanged or grows steadily, while for UFC$_{gen}$, there is often a significant growth on it. For example, when running both algorithms on foodmart, the memory consumption of UFC$_{fast}$ increases steadily as a linear trend, standing at 49.9 MB when the number of transactions is 20,000, with the figure for that being 114.64 MB, when the quantity of transactions is 80,000. By contrast, when applying UFC$_{gen}$, the memory consumption is 1006.88 MB when the transaction number is 20,000. However, the number becomes 2,173 MB when the quantity of transactions is 80,000.

\subsection{Case study on market analysis}

A case study is also evaluated on both proposed algorithms. The real-life dataset is yoochooseBuys, which is a collection of sequences of click events. It represents a period of six months of activities about a big e-commerce business in Europe selling all kinds of stuffs such as garden tools, toys, clothes and electronics. There are 507,746 rows of transactions and 19,102 items in total. Each transaction records the quantity and the price of each item. The frequency threshold is set to range from 0.1\% to 0.25\%, with the threshold of utility starting from 0.05\% to 0.3\% after testing different parameters. Table \ref{tab:yooch_class} shows the detailed classification results.

\begin{table}[h]
	\centering
	\caption{Classification results on yoochooseBuys}
	\setlength{\tabcolsep}{0.4mm}{\begin{tabular}{lcccc}
			\hline
			\textbf{Dataset} & \textbf{\textit{min\_util}} & \textbf{$\#$ \textit{HFHUI}} & \textbf{$\#$ \textit{HFLUI}}& \textbf{$\#$ \textit{LFHUI}} \\ \hline
			\multirow{6}{*}{yoochooseBuys }
			& 0.05 \% &148&68& 348\\
			&0.1 \%&86&130&127\\
			&0.15 \%&62&154&70\\
			&0.20 \%&50&166&40\\
			&0.25 \%&42&174&31\\
			&0.30 \%&28&188&21\\
			\hline
		\end{tabular}
	}
	\label{tab:yooch_class}
\end{table}

\begin{figure*}[!h]
	\centering	
	\includegraphics[height=0.37\textheight,width=0.8\textwidth]{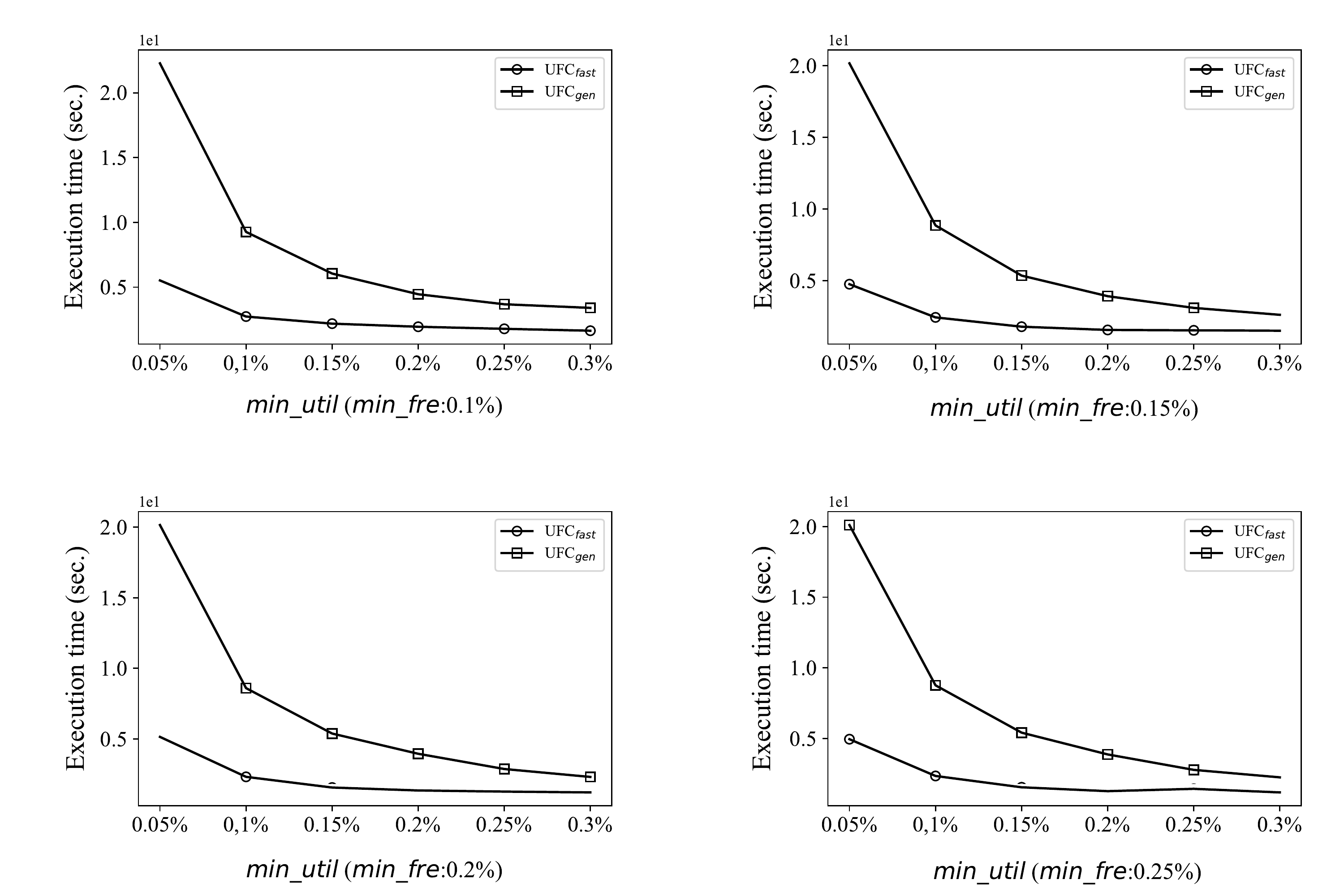}
	\caption{Execution time on yoochooseBuys}
	\label{fig:yooch1}
\end{figure*}

\begin{figure*}[!h]
	\centering
	\includegraphics[height=0.37\textheight,width=0.8\textwidth]{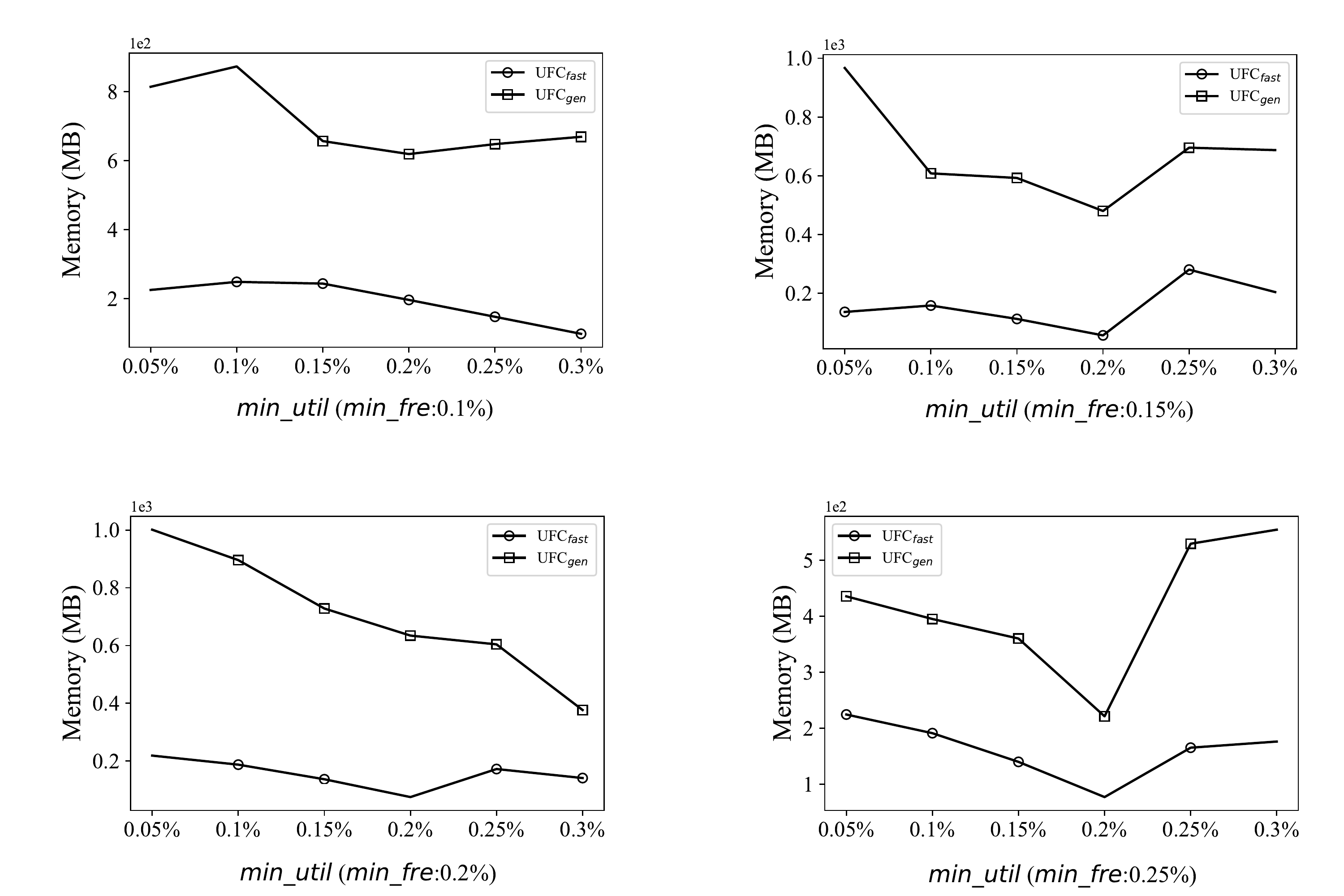}		
	\caption{Memory consumption on yoochooseBuys}
	\label{fig:yooch2}
\end{figure*}

Fig. \ref{fig:yooch1} and Fig. \ref{fig:yooch2} showed the performances of both algorithms with regard to execution time and memory consumption respectively. For the running time, both algorithms have excellent performances on it, presenting a general smooth trend. When the UFC$_{fast}$ is applied, it only takes around 5 s when the utility threshold is 0.05\%, with four different frequency thresholds. As for other combinations of utility and frequency, the running time remains at the level under 3 s, ranging from 1.180 s to 2.731 s. On the other hand, the UFC$_{gen}$  cost more than 20 s when the utility threshold is 0.5\%. However, the time drops sharply when the threshold of utility is transformed into 1\%.  And it shares a similar trend with that for UFC$_{fast}$, and the gap between them is getting smaller. Overall, dramatic differences between both algorithms do not appear on this business dataset from the real world, while both of the proposed methods behave well on it.

\section{Conclusion and Future Studies}
\label{conclusion}
 
In this paper, we studied the problem of pattern classification for smart systems. We used the original properties of both frequency and utility and developed two algorithms, UFC$_{gen}$ and UFC$_{fast}$, which were used to efficiently collect three types of patterns, HFHUIs, HFLUIs, and LFHUIs. For the proposed level-wise-based UFC$_{gen}$ algorithm, it adopted a level-based structure, and a number of new candidates were generated each time when a connection operation was applied. As for UFC$_{fast}$, it utilized a list-based structure, and can reduce the computational time since the database was only scanned twice. Compared with UFC$_{fast}$, UFC$_{gen}$ may not be suitable enough since it required the generation of a large number of candidates  and multiple times of scans for the database during the mining process. Experimental results on real and synthetic datasets shown that both UFC$_{gen}$ and  UFC$_{fast}$ were effective for pattern classification in smart systems. UFC$_{fast}$ had a more significant performance when dealing with different types of datasets under various parameters in terms of both execution time and memory consumption. In practice, UFC$_{fast}$ was more suitable for processing large-scale datasets. 

For future studies, we plan to perform research to further improve the performance of both proposed algorithm. Currently, the number of researches on pattern classification is relatively less, and the method of how to better combine the properties of frequency and utility still requires further discovery. Moreover, studies of pattern classification based on different models in smart systems, such as smart manufacturing, will be performed in the future.

\section{Acknowledgment}

This research was supported in part by the National Natural Science Foundation of China (Grant Nos. 62002136 and 61902079), Guangzhou Basic and Applied Basic Research Foundation (Grant Nos. 202102020277 and 202102020928), Guangdong Basic and Applied Basic Research Foundation (Grant No. 2019B1515120010), Guangdong Key R\&D Plan2020 (Grant No. 2020B0101090002), and National Key R\&D Plan2020 (Grant No. 2020YFB1005600).

\bibliographystyle{ACM-Reference-Format}
\bibliography{paper.bib}


\begin{thebibliography}{50}


\ifx \showCODEN    \undefined \def \showCODEN     #1{\unskip}     \fi
\ifx \showDOI      \undefined \def \showDOI       #1{#1}\fi
\ifx \showISBNx    \undefined \def \showISBNx     #1{\unskip}     \fi
\ifx \showISBNxiii \undefined \def \showISBNxiii  #1{\unskip}     \fi
\ifx \showISSN     \undefined \def \showISSN      #1{\unskip}     \fi
\ifx \showLCCN     \undefined \def \showLCCN      #1{\unskip}     \fi
\ifx \shownote     \undefined \def \shownote      #1{#1}          \fi
\ifx \showarticletitle \undefined \def \showarticletitle #1{#1}   \fi
\ifx \showURL      \undefined \def \showURL       {\relax}        \fi
\providecommand\bibfield[2]{#2}
\providecommand\bibinfo[2]{#2}
\providecommand\natexlab[1]{#1}
\providecommand\showeprint[2][]{arXiv:#2}

\bibitem[\protect\citeauthoryear{Abdelhamid, Ayesh, and Thabtah}{Abdelhamid
  et~al\mbox{.}}{2014}]%
        {abdelhamid2014phishing}
\bibfield{author}{\bibinfo{person}{Neda Abdelhamid}, \bibinfo{person}{Aladdin
  Ayesh}, {and} \bibinfo{person}{Fadi Thabtah}.}
  \bibinfo{year}{2014}\natexlab{}.
\newblock \showarticletitle{Phishing detection based associative classification
  data mining}.
\newblock \bibinfo{journal}{\emph{Expert Systems with Applications}}
  \bibinfo{volume}{41}, \bibinfo{number}{13} (\bibinfo{year}{2014}),
  \bibinfo{pages}{5948--5959}.
\newblock


\bibitem[\protect\citeauthoryear{Abdelhamid and Thabtah}{Abdelhamid and
  Thabtah}{2014}]%
        {abdelhamid2014associative}
\bibfield{author}{\bibinfo{person}{Neda Abdelhamid} {and} \bibinfo{person}{Fadi
  Thabtah}.} \bibinfo{year}{2014}\natexlab{}.
\newblock \showarticletitle{Associative classification approaches: review and
  comparison}.
\newblock \bibinfo{journal}{\emph{Journal of Information \& Knowledge
  Management}} \bibinfo{volume}{13}, \bibinfo{number}{03}
  (\bibinfo{year}{2014}), \bibinfo{pages}{1450027}.
\newblock


\bibitem[\protect\citeauthoryear{Agrawal, Imieli{\'n}ski, and Swami}{Agrawal
  et~al\mbox{.}}{1993}]%
        {agrawal1993mining}
\bibfield{author}{\bibinfo{person}{Rakesh Agrawal}, \bibinfo{person}{Tomasz
  Imieli{\'n}ski}, {and} \bibinfo{person}{Arun Swami}.}
  \bibinfo{year}{1993}\natexlab{}.
\newblock \showarticletitle{Mining association rules between sets of items in
  large databases}. In \bibinfo{booktitle}{\emph{ACM SIGMOD Record}},
  Vol.~\bibinfo{volume}{22}. ACM, \bibinfo{pages}{207--216}.
\newblock


\bibitem[\protect\citeauthoryear{Agrawal, Srikant, et~al\mbox{.}}{Agrawal
  et~al\mbox{.}}{1994}]%
        {agrawal1994fast}
\bibfield{author}{\bibinfo{person}{Rakesh Agrawal},
  \bibinfo{person}{Ramakrishnan Srikant}, {et~al\mbox{.}}}
  \bibinfo{year}{1994}\natexlab{}.
\newblock \showarticletitle{Fast algorithms for mining association rules}. In
  \bibinfo{booktitle}{\emph{The 20th International Conference on Very Large
  Data Bases}}. \bibinfo{pages}{487--499}.
\newblock


\bibitem[\protect\citeauthoryear{Ahmed, Tanbeer, Jeong, and Lee}{Ahmed
  et~al\mbox{.}}{2009}]%
        {ahmed2009efficient}
\bibfield{author}{\bibinfo{person}{Chowdhury~Farhan Ahmed},
  \bibinfo{person}{Syed~Khairuzzaman Tanbeer}, \bibinfo{person}{Byeong~Soo
  Jeong}, {and} \bibinfo{person}{Young~Koo Lee}.}
  \bibinfo{year}{2009}\natexlab{}.
\newblock \showarticletitle{Efficient tree structures for high utility pattern
  mining in incremental databases}.
\newblock \bibinfo{journal}{\emph{IEEE Transactions on Knowledge and Data
  Engineering}} \bibinfo{volume}{21}, \bibinfo{number}{12}
  (\bibinfo{year}{2009}), \bibinfo{pages}{1708--1721}.
\newblock


\bibitem[\protect\citeauthoryear{Baek, Yun, Kim, Kim, Vo, Truong, and
  Deng}{Baek et~al\mbox{.}}{2021}]%
        {baek2021approximate}
\bibfield{author}{\bibinfo{person}{Yoonji Baek}, \bibinfo{person}{Unil Yun},
  \bibinfo{person}{Heonho Kim}, \bibinfo{person}{Jongseong Kim},
  \bibinfo{person}{Bay Vo}, \bibinfo{person}{Tin Truong}, {and}
  \bibinfo{person}{Zhi-Hong Deng}.} \bibinfo{year}{2021}\natexlab{}.
\newblock \showarticletitle{Approximate high utility itemset mining in noisy
  environments}.
\newblock \bibinfo{journal}{\emph{Knowledge-Based Systems}}
  \bibinfo{volume}{212} (\bibinfo{year}{2021}), \bibinfo{pages}{106596}.
\newblock


\bibitem[\protect\citeauthoryear{Brin, Motwani, and Silverstein}{Brin
  et~al\mbox{.}}{1997}]%
        {brin1997beyond}
\bibfield{author}{\bibinfo{person}{Sergey Brin}, \bibinfo{person}{Rajeev
  Motwani}, {and} \bibinfo{person}{Craig Silverstein}.}
  \bibinfo{year}{1997}\natexlab{}.
\newblock \showarticletitle{Beyond market baskets: Generalizing association
  rules to correlations}. In \bibinfo{booktitle}{\emph{ACM SIGMOD Record}},
  Vol.~\bibinfo{volume}{26}. ACM, \bibinfo{pages}{265--276}.
\newblock


\bibitem[\protect\citeauthoryear{Chan, Yang, and Shen}{Chan
  et~al\mbox{.}}{2003}]%
        {chan2003mining}
\bibfield{author}{\bibinfo{person}{Raymond Chan}, \bibinfo{person}{Qiang Yang},
  {and} \bibinfo{person}{Yi~Dong Shen}.} \bibinfo{year}{2003}\natexlab{}.
\newblock \showarticletitle{Mining high utility itemsets}. In
  \bibinfo{booktitle}{\emph{The Third IEEE International Conference on Data
  Mining}}. IEEE, \bibinfo{pages}{19--26}.
\newblock


\bibitem[\protect\citeauthoryear{Chen, Chen, Gan, Qiu, and Ding}{Chen
  et~al\mbox{.}}{2021}]%
        {chen2021discovering}
\bibfield{author}{\bibinfo{person}{Chien-Ming Chen}, \bibinfo{person}{Lili
  Chen}, \bibinfo{person}{Wensheng Gan}, \bibinfo{person}{Lina Qiu}, {and}
  \bibinfo{person}{Weiping Ding}.} \bibinfo{year}{2021}\natexlab{}.
\newblock \showarticletitle{Discovering high utility-occupancy patterns from
  uncertain data}.
\newblock \bibinfo{journal}{\emph{Information Sciences}}  \bibinfo{volume}{546}
  (\bibinfo{year}{2021}), \bibinfo{pages}{1208--1229}.
\newblock


\bibitem[\protect\citeauthoryear{Chen, Han, and Yu}{Chen et~al\mbox{.}}{1996}]%
        {Chen1996Data}
\bibfield{author}{\bibinfo{person}{Ming~Syan Chen}, \bibinfo{person}{Jiawei
  Han}, {and} \bibinfo{person}{Philip~S. Yu}.} \bibinfo{year}{1996}\natexlab{}.
\newblock \showarticletitle{Data mining: an overview from a database
  perspective}.
\newblock \bibinfo{journal}{\emph{IEEE Transactions on Knowledge and Data
  Engineering}} \bibinfo{volume}{8}, \bibinfo{number}{6}
  (\bibinfo{year}{1996}), \bibinfo{pages}{866--883}.
\newblock


\bibitem[\protect\citeauthoryear{Choudhary, Harding, and Tiwari}{Choudhary
  et~al\mbox{.}}{2009}]%
        {choudhary2009data}
\bibfield{author}{\bibinfo{person}{Alok~Kumar Choudhary},
  \bibinfo{person}{Jenny~A Harding}, {and} \bibinfo{person}{Manoj~Kumar
  Tiwari}.} \bibinfo{year}{2009}\natexlab{}.
\newblock \showarticletitle{Data mining in manufacturing: a review based on the
  kind of knowledge}.
\newblock \bibinfo{journal}{\emph{Journal of Intelligent Manufacturing}}
  \bibinfo{volume}{20}, \bibinfo{number}{5} (\bibinfo{year}{2009}),
  \bibinfo{pages}{501--521}.
\newblock


\bibitem[\protect\citeauthoryear{Dogan and Birant}{Dogan and Birant}{2020}]%
        {dogan2020machine}
\bibfield{author}{\bibinfo{person}{Alican Dogan} {and} \bibinfo{person}{Derya
  Birant}.} \bibinfo{year}{2020}\natexlab{}.
\newblock \showarticletitle{Machine learning and data mining in manufacturing}.
\newblock \bibinfo{journal}{\emph{Expert Systems with Applications}}
  (\bibinfo{year}{2020}), \bibinfo{pages}{114060}.
\newblock


\bibitem[\protect\citeauthoryear{Fan, Guo, Chen, Hsu, and Wei}{Fan
  et~al\mbox{.}}{2001}]%
        {fan2001data}
\bibfield{author}{\bibinfo{person}{Chih-Min Fan}, \bibinfo{person}{Ruey-Shan
  Guo}, \bibinfo{person}{Argon Chen}, \bibinfo{person}{Kuo-Ching Hsu}, {and}
  \bibinfo{person}{Chih-Shih Wei}.} \bibinfo{year}{2001}\natexlab{}.
\newblock \showarticletitle{Data mining and fault diagnosis based on wafer
  acceptance test data and in-line manufacturing data}. In
  \bibinfo{booktitle}{\emph{IEEE International Symposium on Semiconductor
  Manufacturing}}. IEEE, \bibinfo{pages}{171--174}.
\newblock


\bibitem[\protect\citeauthoryear{Gan, Lin, Chao, Fournier-Viger, Wang, and
  Yu}{Gan et~al\mbox{.}}{2020a}]%
        {gan2020utility2}
\bibfield{author}{\bibinfo{person}{Wensheng Gan}, \bibinfo{person}{Jerry
  Chun-Wei Lin}, \bibinfo{person}{Han-Chieh Chao}, \bibinfo{person}{Philippe
  Fournier-Viger}, \bibinfo{person}{Xuan Wang}, {and} \bibinfo{person}{Philip~S
  Yu}.} \bibinfo{year}{2020}\natexlab{a}.
\newblock \showarticletitle{Utility-driven mining of trend information for
  intelligent system}.
\newblock \bibinfo{journal}{\emph{ACM Transactions on Management Information
  Systems}} \bibinfo{volume}{11}, \bibinfo{number}{3} (\bibinfo{year}{2020}),
  \bibinfo{pages}{1--28}.
\newblock


\bibitem[\protect\citeauthoryear{Gan, Lin, Chao, Vasilakos, and Yu}{Gan
  et~al\mbox{.}}{2020b}]%
        {gan2020utility}
\bibfield{author}{\bibinfo{person}{Wensheng Gan}, \bibinfo{person}{Jerry
  Chun-Wei Lin}, \bibinfo{person}{Han-Chieh Chao},
  \bibinfo{person}{Athanasios~V Vasilakos}, {and} \bibinfo{person}{Philip~S
  Yu}.} \bibinfo{year}{2020}\natexlab{b}.
\newblock \showarticletitle{Utility-driven data analytics on uncertain data}.
\newblock \bibinfo{journal}{\emph{IEEE Systems Journal}} \bibinfo{volume}{14},
  \bibinfo{number}{3} (\bibinfo{year}{2020}), \bibinfo{pages}{4442--4453}.
\newblock


\bibitem[\protect\citeauthoryear{Gan, Lin, Chao, Wang, and Yu}{Gan
  et~al\mbox{.}}{2018a}]%
        {gan2018privacy}
\bibfield{author}{\bibinfo{person}{Wensheng Gan}, \bibinfo{person}{Jerry
  Chun-Wei Lin}, \bibinfo{person}{Han-Chieh Chao}, \bibinfo{person}{Shyue~Liang
  Wang}, {and} \bibinfo{person}{Philip~S Yu}.}
  \bibinfo{year}{2018}\natexlab{a}.
\newblock \showarticletitle{Privacy preserving utility mining: a survey}. In
  \bibinfo{booktitle}{\emph{IEEE International Conference on Big Data}}. IEEE,
  \bibinfo{pages}{2617--2626}.
\newblock


\bibitem[\protect\citeauthoryear{Gan, Lin, Chao, and Yu}{Gan
  et~al\mbox{.}}{2019}]%
        {gan2019utility}
\bibfield{author}{\bibinfo{person}{Wensheng Gan}, \bibinfo{person}{Jerry
  Chun-Wei Lin}, \bibinfo{person}{Han-Chieh Chao}, {and}
  \bibinfo{person}{Philip~S Yu}.} \bibinfo{year}{2019}\natexlab{}.
\newblock \showarticletitle{Utility-driven mining of high utility episodes}. In
  \bibinfo{booktitle}{\emph{IEEE International Conference on Big Data}}. IEEE,
  \bibinfo{pages}{2644--2653}.
\newblock


\bibitem[\protect\citeauthoryear{Gan, Lin, Chao, and Zhan}{Gan
  et~al\mbox{.}}{2017}]%
        {gan2017data}
\bibfield{author}{\bibinfo{person}{Wensheng Gan}, \bibinfo{person}{Jerry
  Chun-Wei Lin}, \bibinfo{person}{Han-Chieh Chao}, {and}
  \bibinfo{person}{Justin Zhan}.} \bibinfo{year}{2017}\natexlab{}.
\newblock \showarticletitle{Data mining in distributed environment: a survey}.
\newblock \bibinfo{journal}{\emph{Wiley Interdisciplinary Reviews: Data Mining
  and Knowledge Discovery}} \bibinfo{volume}{7}, \bibinfo{number}{6}
  (\bibinfo{year}{2017}), \bibinfo{pages}{e1216}.
\newblock


\bibitem[\protect\citeauthoryear{Gan, Lin, Fournier-Viger, Chao, and
  Fujita}{Gan et~al\mbox{.}}{2018b}]%
        {gan2018extracting}
\bibfield{author}{\bibinfo{person}{Wensheng Gan}, \bibinfo{person}{Jerry
  Chun-Wei Lin}, \bibinfo{person}{Philippe Fournier-Viger},
  \bibinfo{person}{Han-Chieh Chao}, {and} \bibinfo{person}{Hamido Fujita}.}
  \bibinfo{year}{2018}\natexlab{b}.
\newblock \showarticletitle{Extracting non-redundant correlated purchase
  behaviors by utility measure}.
\newblock \bibinfo{journal}{\emph{Knowledge-Based Systems}}
  \bibinfo{volume}{143} (\bibinfo{year}{2018}), \bibinfo{pages}{30--41}.
\newblock


\bibitem[\protect\citeauthoryear{Gan, Lin, Fournier-Viger, Chao, Hong, and
  Fujita}{Gan et~al\mbox{.}}{2018c}]%
        {2gan2018survey}
\bibfield{author}{\bibinfo{person}{Wensheng Gan}, \bibinfo{person}{Jerry
  Chun-Wei Lin}, \bibinfo{person}{Philippe Fournier-Viger},
  \bibinfo{person}{Han-Chieh Chao}, \bibinfo{person}{Tzung-Pei Hong}, {and}
  \bibinfo{person}{Hamido Fujita}.} \bibinfo{year}{2018}\natexlab{c}.
\newblock \showarticletitle{A survey of incremental high-utility itemset
  mining}.
\newblock \bibinfo{journal}{\emph{Wiley Interdisciplinary Reviews: Data Mining
  and Knowledge Discovery}} \bibinfo{volume}{8}, \bibinfo{number}{2}
  (\bibinfo{year}{2018}), \bibinfo{pages}{e1242}.
\newblock


\bibitem[\protect\citeauthoryear{Gan, Lin, Fournier-Viger, Chao, Tseng, and
  Yu}{Gan et~al\mbox{.}}{2021a}]%
        {gan2021survey}
\bibfield{author}{\bibinfo{person}{Wensheng Gan}, \bibinfo{person}{Jerry
  Chun-Wei Lin}, \bibinfo{person}{Philippe Fournier-Viger},
  \bibinfo{person}{Han-Chieh Chao}, \bibinfo{person}{Vincent~S Tseng}, {and}
  \bibinfo{person}{Philip~S Yu}.} \bibinfo{year}{2021}\natexlab{a}.
\newblock \showarticletitle{A survey of utility-oriented pattern mining}.
\newblock \bibinfo{journal}{\emph{IEEE Transactions on Knowledge and Data
  Engineering}} \bibinfo{volume}{33}, \bibinfo{number}{4}
  (\bibinfo{year}{2021}), \bibinfo{pages}{1306--1327}.
\newblock


\bibitem[\protect\citeauthoryear{Gan, Lin, Fournier-Viger, Chao, and Yu}{Gan
  et~al\mbox{.}}{2020c}]%
        {gan2020huopm}
\bibfield{author}{\bibinfo{person}{Wensheng Gan}, \bibinfo{person}{Jerry
  Chun-Wei Lin}, \bibinfo{person}{Philippe Fournier-Viger},
  \bibinfo{person}{Han-Chieh Chao}, {and} \bibinfo{person}{Philip~S Yu}.}
  \bibinfo{year}{2020}\natexlab{c}.
\newblock \showarticletitle{{HUOPM}: High-utility occupancy pattern mining}.
\newblock \bibinfo{journal}{\emph{IEEE Transactions on Cybernetics}}
  \bibinfo{volume}{50}, \bibinfo{number}{3} (\bibinfo{year}{2020}),
  \bibinfo{pages}{1195--1208}.
\newblock


\bibitem[\protect\citeauthoryear{Gan, Lin, Zhang, Chao, Fujita, and Yu}{Gan
  et~al\mbox{.}}{2020d}]%
        {gan2020proum}
\bibfield{author}{\bibinfo{person}{Wensheng Gan}, \bibinfo{person}{Jerry
  Chun-Wei Lin}, \bibinfo{person}{Jiexiong Zhang}, \bibinfo{person}{Han-Chieh
  Chao}, \bibinfo{person}{Hamido Fujita}, {and} \bibinfo{person}{Philip~S Yu}.}
  \bibinfo{year}{2020}\natexlab{d}.
\newblock \showarticletitle{{ProUM}: Projection-based utility mining on
  sequence data}.
\newblock \bibinfo{journal}{\emph{Information Sciences}}  \bibinfo{volume}{513}
  (\bibinfo{year}{2020}), \bibinfo{pages}{222--240}.
\newblock


\bibitem[\protect\citeauthoryear{Gan, Lin, Zhang, Fournier-Viger, Chao, and
  Yu}{Gan et~al\mbox{.}}{2021b}]%
        {gan2021fast}
\bibfield{author}{\bibinfo{person}{Wensheng Gan}, \bibinfo{person}{Jerry
  Chun-Wei Lin}, \bibinfo{person}{Jiexiong Zhang}, \bibinfo{person}{Philippe
  Fournier-Viger}, \bibinfo{person}{Han-Chieh Chao}, {and}
  \bibinfo{person}{Philip~S Yu}.} \bibinfo{year}{2021}\natexlab{b}.
\newblock \showarticletitle{Fast utility mining on sequence data}.
\newblock \bibinfo{journal}{\emph{IEEE Transactions on Cybernetics}}
  \bibinfo{volume}{51}, \bibinfo{number}{2} (\bibinfo{year}{2021}),
  \bibinfo{pages}{487--500}.
\newblock


\bibitem[\protect\citeauthoryear{Han, Kamber, and Pei}{Han
  et~al\mbox{.}}{2011}]%
        {han2011data}
\bibfield{author}{\bibinfo{person}{Jiawei Han}, \bibinfo{person}{Micheline
  Kamber}, {and} \bibinfo{person}{Jian Pei}.} \bibinfo{year}{2011}\natexlab{}.
\newblock \showarticletitle{Data mining concepts and techniques. Third
  edition}.
\newblock \bibinfo{journal}{\emph{The Morgan Kaufmann Series in Data Management
  Systems}} \bibinfo{volume}{5}, \bibinfo{number}{4} (\bibinfo{year}{2011}),
  \bibinfo{pages}{83--124}.
\newblock


\bibitem[\protect\citeauthoryear{Han, Pei, Yin, and Mao}{Han
  et~al\mbox{.}}{2004}]%
        {han2004mining}
\bibfield{author}{\bibinfo{person}{Jiawei Han}, \bibinfo{person}{Jian Pei},
  \bibinfo{person}{Yiwen Yin}, {and} \bibinfo{person}{Runying Mao}.}
  \bibinfo{year}{2004}\natexlab{}.
\newblock \showarticletitle{Mining frequent patterns without candidate
  generation: A frequent-pattern tree approach}.
\newblock \bibinfo{journal}{\emph{Data Mining and Knowledge Discovery}}
  \bibinfo{volume}{8}, \bibinfo{number}{1} (\bibinfo{year}{2004}),
  \bibinfo{pages}{53--87}.
\newblock


\bibitem[\protect\citeauthoryear{Hong, Kuo, and Chi}{Hong
  et~al\mbox{.}}{1999}]%
        {hong1999mining}
\bibfield{author}{\bibinfo{person}{Tzung-Pei Hong}, \bibinfo{person}{Chan~Sheng
  Kuo}, {and} \bibinfo{person}{Sheng~Chai Chi}.}
  \bibinfo{year}{1999}\natexlab{}.
\newblock \showarticletitle{Mining association rules from quantitative data}.
\newblock \bibinfo{journal}{\emph{Intelligent Data Analysis}}
  \bibinfo{volume}{3}, \bibinfo{number}{5} (\bibinfo{year}{1999}),
  \bibinfo{pages}{363--376}.
\newblock


\bibitem[\protect\citeauthoryear{Lin, Fournier-Viger, and Gan}{Lin
  et~al\mbox{.}}{2016a}]%
        {lin2016fhn}
\bibfield{author}{\bibinfo{person}{Jerry Chun-Wei Lin},
  \bibinfo{person}{Philippe Fournier-Viger}, {and} \bibinfo{person}{Wensheng
  Gan}.} \bibinfo{year}{2016}\natexlab{a}.
\newblock \showarticletitle{{FHN}: An efficient algorithm for mining
  high-utility itemsets with negative unit profits}.
\newblock \bibinfo{journal}{\emph{Knowledge-Based Systems}}
  \bibinfo{volume}{111} (\bibinfo{year}{2016}), \bibinfo{pages}{283--298}.
\newblock


\bibitem[\protect\citeauthoryear{Lin, Gan, Fournier-Viger, Hong, and Chao}{Lin
  et~al\mbox{.}}{2017}]%
        {lin2017fdhup}
\bibfield{author}{\bibinfo{person}{Jerry Chun-Wei Lin},
  \bibinfo{person}{Wensheng Gan}, \bibinfo{person}{Philippe Fournier-Viger},
  \bibinfo{person}{Tzung-Pei Hong}, {and} \bibinfo{person}{Han-Chieh Chao}.}
  \bibinfo{year}{2017}\natexlab{}.
\newblock \showarticletitle{{FDHUP}: Fast algorithm for mining discriminative
  high utility patterns}.
\newblock \bibinfo{journal}{\emph{Knowledge and Information Systems}}
  \bibinfo{volume}{51}, \bibinfo{number}{3} (\bibinfo{year}{2017}),
  \bibinfo{pages}{873--909}.
\newblock


\bibitem[\protect\citeauthoryear{Lin, Gan, Fournier-Viger, Hong, and Tseng}{Lin
  et~al\mbox{.}}{2016b}]%
        {lin2016efficient}
\bibfield{author}{\bibinfo{person}{Jerry Chun-Wei Lin},
  \bibinfo{person}{Wensheng Gan}, \bibinfo{person}{Philippe Fournier-Viger},
  \bibinfo{person}{Tzung-Pei Hong}, {and} \bibinfo{person}{Vincent~S Tseng}.}
  \bibinfo{year}{2016}\natexlab{b}.
\newblock \showarticletitle{Efficient algorithms for mining high-utility
  itemsets in uncertain databases}.
\newblock \bibinfo{journal}{\emph{Knowledge-Based Systems}}
  \bibinfo{volume}{96} (\bibinfo{year}{2016}), \bibinfo{pages}{171--187}.
\newblock


\bibitem[\protect\citeauthoryear{Lin, Gan, Hong, and Tseng}{Lin
  et~al\mbox{.}}{2015}]%
        {lin2015efficient}
\bibfield{author}{\bibinfo{person}{Jerry Chun-Wei Lin},
  \bibinfo{person}{Wensheng Gan}, \bibinfo{person}{Tzung-Pei Hong}, {and}
  \bibinfo{person}{Vincent~S Tseng}.} \bibinfo{year}{2015}\natexlab{}.
\newblock \showarticletitle{Efficient algorithms for mining up-to-date
  high-utility patterns}.
\newblock \bibinfo{journal}{\emph{Advanced Engineering Informatics}}
  \bibinfo{volume}{29}, \bibinfo{number}{3} (\bibinfo{year}{2015}),
  \bibinfo{pages}{648--661}.
\newblock


\bibitem[\protect\citeauthoryear{Lin, Gan, Wu, Chen, and Chen}{Lin
  et~al\mbox{.}}{2021}]%
        {lin2021joint}
\bibfield{author}{\bibinfo{person}{Qi Lin}, \bibinfo{person}{Wensheng Gan},
  \bibinfo{person}{Yongdong Wu}, \bibinfo{person}{Jiahui Chen}, {and}
  \bibinfo{person}{Chien-Ming Chen}.} \bibinfo{year}{2021}\natexlab{}.
\newblock \showarticletitle{Joint utility and frequency for pattern
  classification}. In \bibinfo{booktitle}{\emph{IEEE International Conference
  on Big Data}}. IEEE, \bibinfo{pages}{5524--5533}.
\newblock


\bibitem[\protect\citeauthoryear{Liu and Qu}{Liu and Qu}{2012}]%
        {liu2012mining}
\bibfield{author}{\bibinfo{person}{Mengchi Liu} {and} \bibinfo{person}{Junfeng
  Qu}.} \bibinfo{year}{2012}\natexlab{}.
\newblock \showarticletitle{Mining high utility itemsets without candidate
  generation}. In \bibinfo{booktitle}{\emph{The 21st ACM International
  Conference on Information and Knowledge Management}}. ACM,
  \bibinfo{pages}{55--64}.
\newblock


\bibitem[\protect\citeauthoryear{Liu, Liao, and Choudhary}{Liu
  et~al\mbox{.}}{2005}]%
        {liu2005two}
\bibfield{author}{\bibinfo{person}{Ying Liu}, \bibinfo{person}{Wei~keng Liao},
  {and} \bibinfo{person}{Alok Choudhary}.} \bibinfo{year}{2005}\natexlab{}.
\newblock \showarticletitle{A two-phase algorithm for fast discovery of high
  utility itemsets}. In \bibinfo{booktitle}{\emph{The Pacific-Asia Conference
  on Knowledge Discovery and Data Mining}}. Springer,
  \bibinfo{pages}{689--695}.
\newblock


\bibitem[\protect\citeauthoryear{Luna, Fournier-Viger, and Ventura}{Luna
  et~al\mbox{.}}{2019}]%
        {luna2019frequent}
\bibfield{author}{\bibinfo{person}{Jos{\'e}~Mar{\'\i}a Luna},
  \bibinfo{person}{Philippe Fournier-Viger}, {and}
  \bibinfo{person}{Sebasti{\'a}n Ventura}.} \bibinfo{year}{2019}\natexlab{}.
\newblock \showarticletitle{Frequent itemset mining: A 25 years review}.
\newblock \bibinfo{journal}{\emph{Wiley Interdisciplinary Reviews: Data Mining
  and Knowledge Discovery}} \bibinfo{volume}{9}, \bibinfo{number}{6}
  (\bibinfo{year}{2019}), \bibinfo{pages}{e1329}.
\newblock


\bibitem[\protect\citeauthoryear{Nakata, Orihara, Mizuoka, and Takagi}{Nakata
  et~al\mbox{.}}{2017}]%
        {nakata2017comprehensive}
\bibfield{author}{\bibinfo{person}{Kouta Nakata}, \bibinfo{person}{Ryohei
  Orihara}, \bibinfo{person}{Yoshiaki Mizuoka}, {and} \bibinfo{person}{Kentaro
  Takagi}.} \bibinfo{year}{2017}\natexlab{}.
\newblock \showarticletitle{A comprehensive big-data-based monitoring system
  for yield enhancement in semiconductor manufacturing}.
\newblock \bibinfo{journal}{\emph{IEEE Transactions on Semiconductor
  Manufacturing}} \bibinfo{volume}{30}, \bibinfo{number}{4}
  (\bibinfo{year}{2017}), \bibinfo{pages}{339--344}.
\newblock


\bibitem[\protect\citeauthoryear{Nguyen, Nguyen, Nguyen, Vo, Fournier-Viger,
  and Tseng}{Nguyen et~al\mbox{.}}{2019}]%
        {nguyen2019mining}
\bibfield{author}{\bibinfo{person}{Loan~TT Nguyen}, \bibinfo{person}{Phuc
  Nguyen}, \bibinfo{person}{Trinh~DD Nguyen}, \bibinfo{person}{Bay Vo},
  \bibinfo{person}{Philippe Fournier-Viger}, {and} \bibinfo{person}{Vincent~S
  Tseng}.} \bibinfo{year}{2019}\natexlab{}.
\newblock \showarticletitle{Mining high-utility itemsets in dynamic profit
  databases}.
\newblock \bibinfo{journal}{\emph{Knowledge-Based Systems}}
  \bibinfo{volume}{175} (\bibinfo{year}{2019}), \bibinfo{pages}{130--144}.
\newblock


\bibitem[\protect\citeauthoryear{Nguyen, Vo, Hong, and Thanh}{Nguyen
  et~al\mbox{.}}{2012}]%
        {nguyen2012classification}
\bibfield{author}{\bibinfo{person}{Loan~TT Nguyen}, \bibinfo{person}{Bay Vo},
  \bibinfo{person}{Tzung-Pei Hong}, {and} \bibinfo{person}{Hoang~Chi Thanh}.}
  \bibinfo{year}{2012}\natexlab{}.
\newblock \showarticletitle{Classification based on association rules: A
  lattice-based approach}.
\newblock \bibinfo{journal}{\emph{Expert Systems with Applications}}
  \bibinfo{volume}{39}, \bibinfo{number}{13} (\bibinfo{year}{2012}),
  \bibinfo{pages}{11357--11366}.
\newblock


\bibitem[\protect\citeauthoryear{Pei, Han, Lu, Nishio, Tang, and Yang}{Pei
  et~al\mbox{.}}{2001}]%
        {pei2001h}
\bibfield{author}{\bibinfo{person}{Jian Pei}, \bibinfo{person}{Jiawei Han},
  \bibinfo{person}{Hongjun Lu}, \bibinfo{person}{Shojiro Nishio},
  \bibinfo{person}{Shiwei Tang}, {and} \bibinfo{person}{Dongqing Yang}.}
  \bibinfo{year}{2001}\natexlab{}.
\newblock \showarticletitle{H-{M}ine: Hyper-structure mining of frequent
  patterns in large databases}. In \bibinfo{booktitle}{\emph{IEEE International
  Conference on Data Mining}}. IEEE, \bibinfo{pages}{441--448}.
\newblock


\bibitem[\protect\citeauthoryear{Shankar, Babu, Purusothaman, and
  Jayanthi}{Shankar et~al\mbox{.}}{2009}]%
        {shankar2009fast}
\bibfield{author}{\bibinfo{person}{S Shankar}, \bibinfo{person}{Nishanth Babu},
  \bibinfo{person}{T Purusothaman}, {and} \bibinfo{person}{S Jayanthi}.}
  \bibinfo{year}{2009}\natexlab{}.
\newblock \showarticletitle{A fast algorithm for mining high utility itemsets}.
  In \bibinfo{booktitle}{\emph{IEEE International Advance Computing
  Conference}}. IEEE, \bibinfo{pages}{1459--1464}.
\newblock


\bibitem[\protect\citeauthoryear{Shao, Yin, Liu, and Cao}{Shao
  et~al\mbox{.}}{2015}]%
        {shao2015mining}
\bibfield{author}{\bibinfo{person}{Jingyu Shao}, \bibinfo{person}{Junfu Yin},
  \bibinfo{person}{Wei Liu}, {and} \bibinfo{person}{Longbing Cao}.}
  \bibinfo{year}{2015}\natexlab{}.
\newblock \showarticletitle{Mining actionable combined patterns of high utility
  and frequency}. In \bibinfo{booktitle}{\emph{IEEE International Conference on
  Data Science and Advanced Analytics}}. IEEE, \bibinfo{pages}{1--10}.
\newblock


\bibitem[\protect\citeauthoryear{Thabtah}{Thabtah}{2007}]%
        {thabtah2007review}
\bibfield{author}{\bibinfo{person}{Fadi~Abdeljaber Thabtah}.}
  \bibinfo{year}{2007}\natexlab{}.
\newblock \showarticletitle{A review of associative classification mining}.
\newblock \bibinfo{journal}{\emph{Knowledge Engineering Review}}
  \bibinfo{volume}{22}, \bibinfo{number}{1} (\bibinfo{year}{2007}),
  \bibinfo{pages}{37--65}.
\newblock


\bibitem[\protect\citeauthoryear{Thabtah, Cowling, and Peng}{Thabtah
  et~al\mbox{.}}{2004}]%
        {thabtah2004mmac}
\bibfield{author}{\bibinfo{person}{Fadi~A Thabtah}, \bibinfo{person}{Peter
  Cowling}, {and} \bibinfo{person}{Yonghong Peng}.}
  \bibinfo{year}{2004}\natexlab{}.
\newblock \showarticletitle{{MMAC}: A new multi-class, multi-label associative
  classification approach}. In \bibinfo{booktitle}{\emph{Fourth IEEE
  International Conference on Data Mining}}. IEEE, \bibinfo{pages}{217--224}.
\newblock


\bibitem[\protect\citeauthoryear{Tseng, Wu, Fournier-Viger, and Yu}{Tseng
  et~al\mbox{.}}{2015}]%
        {tseng2015efficient}
\bibfield{author}{\bibinfo{person}{Vincent~S Tseng}, \bibinfo{person}{Cheng-Wei
  Wu}, \bibinfo{person}{Philippe Fournier-Viger}, {and}
  \bibinfo{person}{Philip~S Yu}.} \bibinfo{year}{2015}\natexlab{}.
\newblock \showarticletitle{Efficient algorithms for mining top-$k$ high
  utility itemsets}.
\newblock \bibinfo{journal}{\emph{IEEE Transactions on Knowledge and Data
  Engineering}} \bibinfo{volume}{28}, \bibinfo{number}{1}
  (\bibinfo{year}{2015}), \bibinfo{pages}{54--67}.
\newblock


\bibitem[\protect\citeauthoryear{Tseng, Wu, Shie, and Yu}{Tseng
  et~al\mbox{.}}{2010}]%
        {tseng2010up}
\bibfield{author}{\bibinfo{person}{Vincent~S Tseng}, \bibinfo{person}{Cheng~Wei
  Wu}, \bibinfo{person}{Bai~En Shie}, {and} \bibinfo{person}{Philip~S Yu}.}
  \bibinfo{year}{2010}\natexlab{}.
\newblock \showarticletitle{{UP-G}rowth: an efficient algorithm for high
  utility itemset mining}. In \bibinfo{booktitle}{\emph{The 16th ACM SIGKDD
  International Conference on Knowledge Discovery and Data Mining}}. ACM,
  \bibinfo{pages}{253--262}.
\newblock


\bibitem[\protect\citeauthoryear{Wang, Liu, Zhou, Shi, and Zhu}{Wang
  et~al\mbox{.}}{2007}]%
        {wang2007pushing}
\bibfield{author}{\bibinfo{person}{Jing Wang}, \bibinfo{person}{Ying Liu},
  \bibinfo{person}{Lin Zhou}, \bibinfo{person}{Yong Shi}, {and}
  \bibinfo{person}{Xingquan Zhu}.} \bibinfo{year}{2007}\natexlab{}.
\newblock \showarticletitle{Pushing frequency constraint to utility mining
  model}. In \bibinfo{booktitle}{\emph{International Conference on
  Computational Science}}. Springer, \bibinfo{pages}{685--692}.
\newblock


\bibitem[\protect\citeauthoryear{Xie and Yu}{Xie and Yu}{2010}]%
        {xie2010max}
\bibfield{author}{\bibinfo{person}{Yan Xie} {and} \bibinfo{person}{Philip~S
  Yu}.} \bibinfo{year}{2010}\natexlab{}.
\newblock \showarticletitle{Max-clique: A top-down graph-based approach to
  frequent pattern mining}. In \bibinfo{booktitle}{\emph{IEEE International
  Conference on Data Mining}}. IEEE, \bibinfo{pages}{1139--1144}.
\newblock


\bibitem[\protect\citeauthoryear{Zaki}{Zaki}{2000}]%
        {zaki2000scalable}
\bibfield{author}{\bibinfo{person}{Mohammed~Javeed Zaki}.}
  \bibinfo{year}{2000}\natexlab{}.
\newblock \showarticletitle{Scalable algorithms for association mining}.
\newblock \bibinfo{journal}{\emph{IEEE Transactions on Knowledge and Data
  Engineering}} \bibinfo{volume}{12}, \bibinfo{number}{3}
  (\bibinfo{year}{2000}), \bibinfo{pages}{372--390}.
\newblock


\bibitem[\protect\citeauthoryear{Zhang, Du, Gan, and Yu}{Zhang
  et~al\mbox{.}}{2021a}]%
        {zhang2021tkus}
\bibfield{author}{\bibinfo{person}{Chunkai Zhang}, \bibinfo{person}{Zilin Du},
  \bibinfo{person}{Wensheng Gan}, {and} \bibinfo{person}{Philip~S Yu}.}
  \bibinfo{year}{2021}\natexlab{a}.
\newblock \showarticletitle{{TKUS}: Mining top-$k$ high utility sequential
  patterns}.
\newblock \bibinfo{journal}{\emph{Information Sciences}}  \bibinfo{volume}{570}
  (\bibinfo{year}{2021}), \bibinfo{pages}{342--359}.
\newblock


\bibitem[\protect\citeauthoryear{Zhang, Du, Yang, Gan, and Yu}{Zhang
  et~al\mbox{.}}{2021b}]%
        {zhang2021shelf}
\bibfield{author}{\bibinfo{person}{Chunkai Zhang}, \bibinfo{person}{Zilin Du},
  \bibinfo{person}{Yuting Yang}, \bibinfo{person}{Wensheng Gan}, {and}
  \bibinfo{person}{Philip~S Yu}.} \bibinfo{year}{2021}\natexlab{b}.
\newblock \showarticletitle{On-shelf utility mining of sequence data}.
\newblock \bibinfo{journal}{\emph{ACM Transactions on Knowledge Discovery from
  Data}} \bibinfo{volume}{16}, \bibinfo{number}{2} (\bibinfo{year}{2021}),
  \bibinfo{pages}{1--31}.
\newblock


\end{thebibliography}
\end{document}